\pdfoutput=1

\documentclass[11pt]{article}

\usepackage[final]{acl}

\usepackage{times}
\usepackage{latexsym}
\usepackage{booktabs}

\usepackage[T1]{fontenc}

\usepackage[utf8]{inputenc}

\usepackage{microtype}
\usepackage{comment}
\usepackage{xspace}
\usepackage{multirow}
\usepackage{multicol}
\usepackage{graphicx}
\usepackage{booktabs}
\usepackage{todonotes}
\usepackage{tablefootnote}
\usepackage{soul}
\usepackage{subfigure}
\usepackage{listings}
\usepackage{pgfplots}
\usepackage{amssymb}

\usepackage{inconsolata}
\usepackage[normalem]{ulem}
\definecolor{codegreen}{rgb}{0,0.6,0}
\definecolor{codegray}{rgb}{0.5,0.5,0.5}
\definecolor{codepurple}{rgb}{0.58,0,0.82}

\lstdefinestyle{mystyle}{
  frame=single,
  basicstyle=\ttfamily\footnotesize,
  backgroundcolor=\color{backcolour}, commentstyle=\color{codegreen},
  commentstyle=\color{darkgreen}\slshape,
  keywordstyle=\color{blue},
  stringstyle=\color{darkred},
  numberstyle=\tiny\color{codegray},
  emphstyle=\color{pink}\underbar,
  morekeywords={Verify, Question},
  escapeinside={(*@}{@*)},
  breakatwhitespace=false,         
  breaklines=true,                 
  captionpos=b,                    
  keepspaces=true,                    
  numbersep=5pt,                  
  showspaces=false,                
  showstringspaces=false,
  showtabs=false,                  
  tabsize=2
}
\useunder{\uline}{\ul}{}
\newcommand{\dataset}{\textsc{DataTales}\xspace}

\definecolor{kmycolor}{rgb}{0.858, 0.188, 0.478}

\title{\dataset{}: A Benchmark for Real-World Intelligent Data Narration}

\author{
    Yajing Yang$^{1,2}$, Qian Liu$^{3}$, Min-Yen Kan$^{1}$
    \vspace{2mm} \\
    \textsuperscript{1}National University of Singapore\;\;  \textsuperscript{2}Rio Tinto\;\;
    \textsuperscript{3}Sea AI Lab\\
    {\texttt{\small yajing.yang@u.nus.edu\;;liuqian@sea.com\;; kanmy@comp.nus.edu.sg}}
}

\begin{document}
\maketitle
\begin{abstract}
We introduce \dataset{}, a novel benchmark designed to assess the proficiency of language models in data narration, a task crucial for transforming complex tabular data into accessible narratives. Existing benchmarks often fall short in capturing the requisite analytical complexity for practical applications. \dataset{} addresses this gap by offering 4.9k financial reports paired with corresponding market data, showcasing the demand for models to create clear narratives and analyze large datasets while understanding specialized terminology in the field. Our findings highlight the significant challenge that language models face in achieving the necessary precision and analytical depth for proficient data narration, suggesting promising avenues for future model development and evaluation methodologies. The data and code are available at \url{https://github.com/yajingyang/DataTales/}.

\end{abstract}


\section{Introduction}

\begin{figure}[tb]
    \centering
    \includegraphics[width=1\columnwidth]{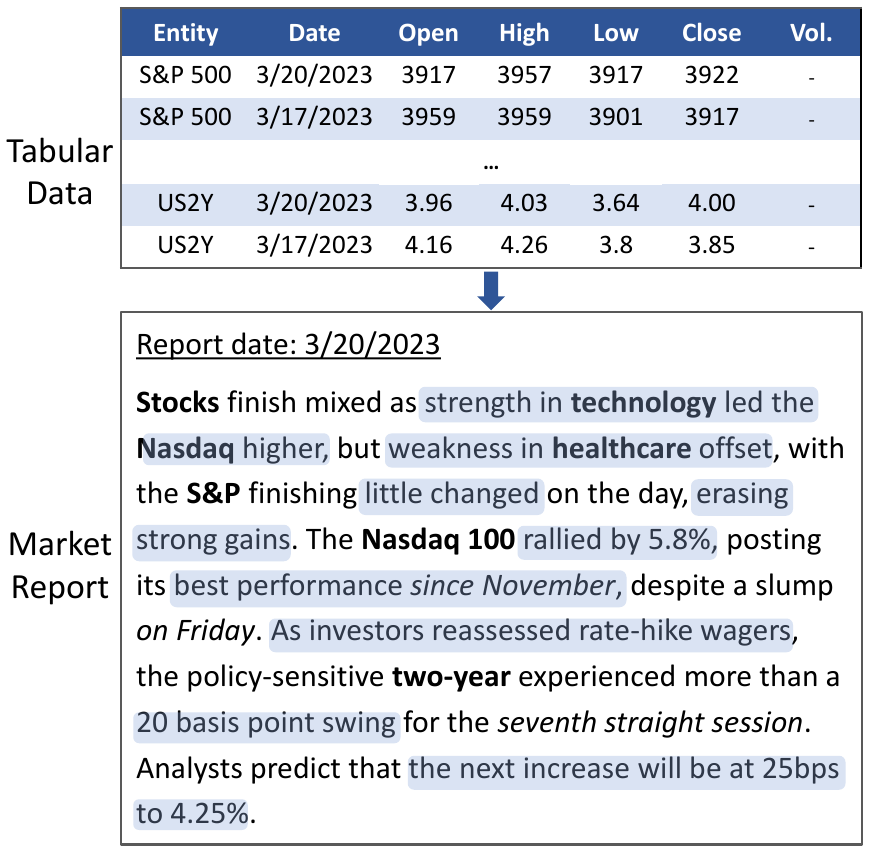}
    \caption{\dataset{} example featuring a report and tabular data on 28 equity market entities, with 7 columns. Bolded text denotes the six mentioned entities. Historical references cover periods of months (``since November''), day of the week (``on Friday''), and days (``seventh straight session''), as italicised. Blue text describes analyses, such as trend (``a 20 basis point swing''), causal (``investors reassessed rate-hike wagers''), and predictive analysis (``next increase predicted at 25bps to 4.25\%'').
    }
    \label{fig:data-example}
     \vspace{-0.5cm}  
\end{figure}



Data narration, the process of transforming intricate data into compelling narratives~\citep{Dourish_Gómez_2018}, plays a critical role in shaping business decision-making.
By distilling vast amounts of information into digestible narratives, it empowers executives with clear and actionable insights~\citep{dykes_chapter_2019, el_outa_towards_2020}.
Moreover, it fosters accessibility to valuable information, reaching a wider audience.
However, traditional manual approaches are burdened by both time constraints and the potential for inaccuracies.
Consequently, there has been a 
longstanding anticipation for models capable of autonomously extracting meaningful insights from data~\citep{DBLP:journals/corr/DemiralpHPP17, ding_quickinsights_2019}.
The rise of large language models (LLMs), such as GPT-3~\citep{gpt3paper} and Llama~\citep{llama_paper}, signifies a beacon of hope within the field.
These models demonstrate extraordinary capabilities, evidenced by their increasing utilization in advanced data analyses~\citep{openagents_paper}.
Empirical evidence highlights that LLMs are effective in reasoning and analytical tasks, achieving performance comparable to or exceeding humans in certain areas~\citep{openai2023gpt4, anthropic2024claude}.
Their ability to understand and generate fluent natural language sentences suggests their potential for data narration tasks.
This leads to an important research question: \textit{Can LLMs achieve proficiency on data narration?}

However, assessing the proficiency of LLMs in data narration is hindered by the limitations of existing benchmarks.
Though related to data-to-text, data narration's complexity surpasses current data-to-text tasks which focus on basic information transformation.
Datasets such as RotoWire~\citep{wiseman_challenges_2017}, WikiBio~\citep{liu_table--text_2018}, and ToTTo~\citep{parikh_totto_2020} translate the original information (e.g., table cells) into another format (e.g., descriptions), without incorporating complex analytical operations.
In contrast, data narration involves a deeper analysis to craft narratives around key insights, as illustrated in the equity market report in Figure~\ref{fig:data-example}. 
The report describes stock index data, analyzes trends, explores causes, and makes predictions.

We introduce \dataset{}, a benchmark comprising 4.9k financial market reports paired with corresponding tabular data, designed to address the current benchmark challenges in data narration.
\dataset{} reports are sourced from diverse outlets and paired with comprehensive financial ticker data (Figure~\ref{fig:data_collection}),
emphasizing in-depth analysis over an extensive input data narrated with professional language, mirroring real-world data narration challenges (Figure~\ref{fig:data-example}).
Our analyses on \dataset{} highlight its support for complex analytical tasks, the importance of domain-specific terminology, and the necessity of selecting from extensive input data to accurately replicate nuanced reports. 
Benchmarking state-of-the-art models on \dataset{} in zero-shot and fine-tuning settings reveals their struggle to achieve the required accuracy and analytical depth, emphasizing the need for models with advanced reasoning over extensive data. Our analyses also expose a significant gap in current automated evaluations for assessing data narratives quality.



\section{Related Work}

\paragraph{Data-to-Text Generation.} Datasets like RotoWire~\citep{wiseman_challenges_2017}, WikiBio~\citep{liu_table--text_2018} and ToTTo~\citep{parikh_totto_2020} 
convert data to text in open domains, providing coherent data descriptions but lacking substantial reasoning crucial for generating insightful financial narratives. This limitation is also observed in domain-specific datasets (WeatherGov~\citep{Liang_Jordan_Klein_2009}, E2E~\citep{Novikova_Lemon_Rieser_2016}, MLB~\citep{wiseman_challenges_2017}, and Dart~\citep{Nan_dart_2021}) and those emphasizing short inputs and limited analysis types (LogicNLG~\citep{Chen_Chen_Su_Chen_Wang_2020}, NumericNLG~\citep{suadaa_towards_2021}, and SciGen~\citep{moosavi_scigen_2021}, 
such as  simple arithmetic and causal analysis. These characteristics contrast with the extensive complex reasoning required for proficiently narrating extensive data. 

\paragraph{Table Insight Generation.} 
PivotTable~\citep{zhou_table2analysis_2020} and AnaMeta~\citep{he_anameta_2023} are datasets designed to transform table data into structured insights, with PivotTable focusing on data aggregation and reasoning, and AnaMeta enhancing field semantics with derived supervision labels. Methodologically,
Foresight~\citep{DBLP:journals/corr/DemiralpHPP17}, 
Voder~\citep{Srinivasan_Drucker_Endert_Stasko_2019}, DataShot~\citep{Wang_Sun_Zhang_Cui_Xu_Ma_Zhang_2020}, Table2analysis~\citep{zhou_table2analysis_2020}, and 
Calliope~\citep{Shi_Xu_Sun_Shi_Cao_2021} propose insight classification taxonomies and utilize recommendation assessment metrics.
Contrasting against their primary focus on visual representations. Our work emphasizes textual narratives to meet the data narration demand, 

\paragraph{Financial NLP.} Financial NLP tasks encompass a wide spectrum, ranging from fraud detection, which aims to identify irregular activities~\citep{Boulieris_Pavlopoulos_Xenos_Vassalos_2023}, to sentiment analysis, which assesses market sentiment through nuanced language interpretation ~\citep{Malo_Sinha_Korhonen_Wallenius_Takala_2014,Atzeni_Dridi_2017,Maia_Handschuh_2018}. 
Question answering tasks, such as FiQA~\citep{Maia_Handschuh_2018}, TAT-QA \citep{Zhu_Lei_Huang_Wang_Zhang_Lv_Feng_Chua_2021}, FinQA~\citep{chen_finqa_2022} and ConvFinQA~\citep{chen_convfinqa_2022},  further amplify the complexity by requiring comprehensive financial data synthesis.
Despite illustrating significant advancements in reasoning complexity, these tasks often lack the analytical depth required for data narration. To the best of our knowledge, we are the first to release a data narration-tailored benchmark.

\paragraph{News Narration.} News narration focuses on extracting narratives from unstructured text, such as news articles or social media posts~\citep{Santana_Campos_Amorim_Jorge_Silvano_Nunes_2023, Keith_Norambuena_Mitra_North_2023}. Generating news narratives requires the identification of events and participants, and linking them by their temporal or spatial information~\citep{Chieu_Lee_2004, Nallapati_Feng_Peng_Allan_2004, Chen_Chen_2012, Wei_Lee_Chiang_Chen_Yang_2014, Chen_Niu_Fu_2015}. In contrast, data narration involves identifying patterns and trends from structured data, 
which often requires complex reasoning over multiple data points.

\section{The \dataset{} Benchmark}



We outline our data collection procedure employed for compiling \dataset{}. Subsequently, we conduct a comprehensive analysis to underscore its unique contributions.

\subsection{Dataset collection}

\begin{figure}[tb]
    \centering
    \includegraphics[width=7.5cm]{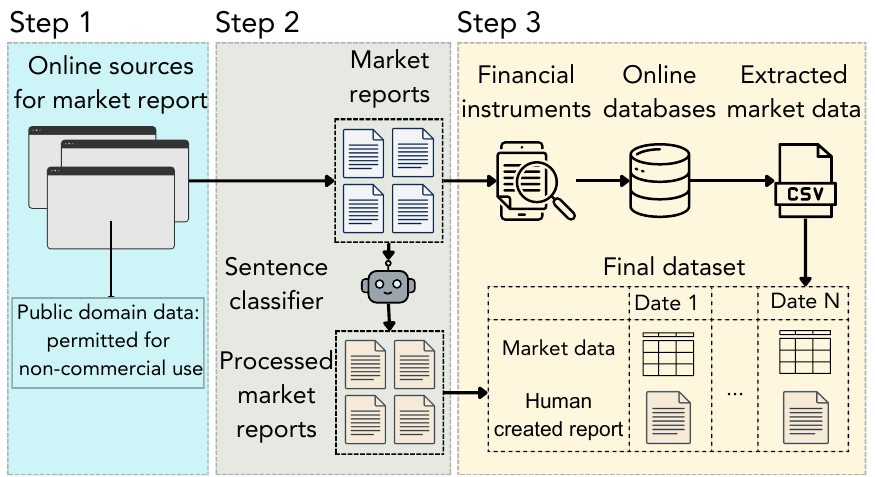}
    \caption{Steps in collecting \dataset{}. 
    }
    \label{fig:data_collection}
    \vspace{-0.5cm}  
\end{figure}

The creation process involves three key steps to curate a dataset for data narration (Figure~\ref{fig:data_collection}).

\paragraph{Step 1: Market Report Collection.} We select online sources that publish daily market reports covering a wide range of sectors (equity, treasury, currency, commodities) with significant analytical depth, including causal analysis, trend analysis, and predictions. The chosen platforms for this purpose are Investrade, Totalfarmmarketing, VT Markets, and LeapRate\footnote{\url{https://www.totalfarmmarketing.com/}, \url{https://www.leaprate.com/}, \url{https://www.vtmarkets.com/}, \url{https://www.investrade.com/}}. From these sources, we compile a dataset of 4.9k reports, ensuring comprehensive market coverage and analytical rigor.

\paragraph{Step 2: Sentence Classification.}
We enhance the dataset by focusing on narratives grounded in tabular data. We employ ChatGPT with in-context learning for sentence-level classification, categorizing sentences into Market Movements, Market Context, External Events and Influence, and Prediction and Suggestion, based on the main type of information they convey. (details in Appendix~\ref{sec:appendix-sentence_classification}). 
Retaining only Market Movements and Predictions sentences ensures that the content is derived from tabular data.  This reduces the report to 54.4\% of its original length on average, and focuses each report on data-driven insights.

\paragraph{Step 3: Data Extraction and Alignment.}
We obtain the corresponding tabular data by identifying the commonly described financial instruments, and extracting data from Yahoo! Finance\footnote{Via use of the yfinance library: \url{https://github.com/ranaroussi/yfinance}}, CME\footnote{\url{https://www.cmegroup.com/}}, Investing.com\footnote{\url{https://www.investing.com/}}, WSJ\footnote{\url{https://www.wsj.com/}}, and Barchart\footnote{\url{https://www.barchart.com/}}. Our manual verification process involves sampling reports across different markets and publishers, directly comparing reported values with downloaded data, and cross-checking derived calculations. We account for potential timing differences by examining adjacent days' data when discrepancies arise, and seek out alternative sources when necessary to ensure data accuracy across various market data providers.\\

Table~\ref{tab:dataset_comparison} illustrates \dataset{}'s unique position among data-to-text and financial NLP benchmarks, offering a combination of large input sizes and advanced analytics capabilities.

\begin{table*}[tb]
\centering

\resizebox{\textwidth}{!}{%
\begin{tabular}{@{}ccccccc@{}}
\toprule
\textbf{Dataset} & \textbf{Task} & \textbf{Pairs} & \textbf{Domain} & \textbf{Input Data Size} & \textbf{Avg. Output Len.} & \textbf{Advanced Anlaysis} \\ \midrule
QuickInsight & Visual Recommendation & 486 & Open & Large & - & None \\
TAT-QA & Question Answering & 16.5K & Finance & Moderate & - & Causal Relation, Trend \\
FinQA & Question Answering & 8.3K & Finance & Moderate & - & Causal Relation, Trend \\
ToTTo & Data-to-Text & 136K & Open & Small & 17 & None \\
RotoWire & Data-to-Text & 11K & Sports & Moderate & 337 & None \\
SciGen & Data-to-Text & 1.3K & Computing & Small & 116 & Causal Relation \\
\dataset{} & Data-to-Text & 4.9K & Finance & Large & 108 & Causal Relation, Trend, Prediction \\ \bottomrule
\end{tabular}%

}
\caption{Comparison of \dataset{} against QuickInsight~\citep{ding_quickinsights_2019}, TAT-QA~\citep{Zhu_Lei_Huang_Wang_Zhang_Lv_Feng_Chua_2021} FinQA~\citep{chen_finqa_2022},  ToTTo~\citep{parikh_totto_2020}, RotoWire~\citep{wiseman_challenges_2017} and SciGen~\citep{moosavi_scigen_2021}, 
presenting statistics related to the task, number of input-output pairs, domain, size of tabular data per input, average number of tokens in target text, and advanced analysis types involved.}
\label{tab:dataset_comparison}
\end{table*}




\begin{table*}[tb]
\centering
\resizebox{14cm}{!}{%
\begin{tabular}{@{}llcl@{}}
\toprule
\textbf{Operation} & \textbf{Category} & \textbf{\% Sent} & \textbf{Example} \\ \midrule
Lookup & Simple & 83.39 & April live cattle future settled at \$19.93. \\
Comparison & Basic Quantitative & 62.03 & Brent crude oil closed higher today. \\
Causal Analysis & Advance Analytical & 38.31 & This price move is driven by fundamental factors. \\
Trend Analysis & Advance Analytical & 31.53 & The US Dollar experienced a continuous decline. \\
Subtraction & Basic Quantitative & 19.66 & The barrel trade added 4.75c to \$1.8875/lb. \\
Predictive Analysis & Advance Analytical & 14.24 & The euro is likely to stay under pressure. \\
Rate of Change & Basic Quantitative & 12.54 & The Dow Jones Industrial Average fell 0.65\%. \\ \bottomrule
\end{tabular}%
}
\caption{Reasoning operations used in generating market reports, based on the manual analysis results of 295 sentences from 20 reports. Details include the category of complexity,  
operation prevalence (in \%age), and examples.}
\label{tab:reasoning-operations}
\end{table*}

\subsection{Analytical Operations Analysis}
\begin{figure}[tb]
    \includegraphics[width=7.8cm]{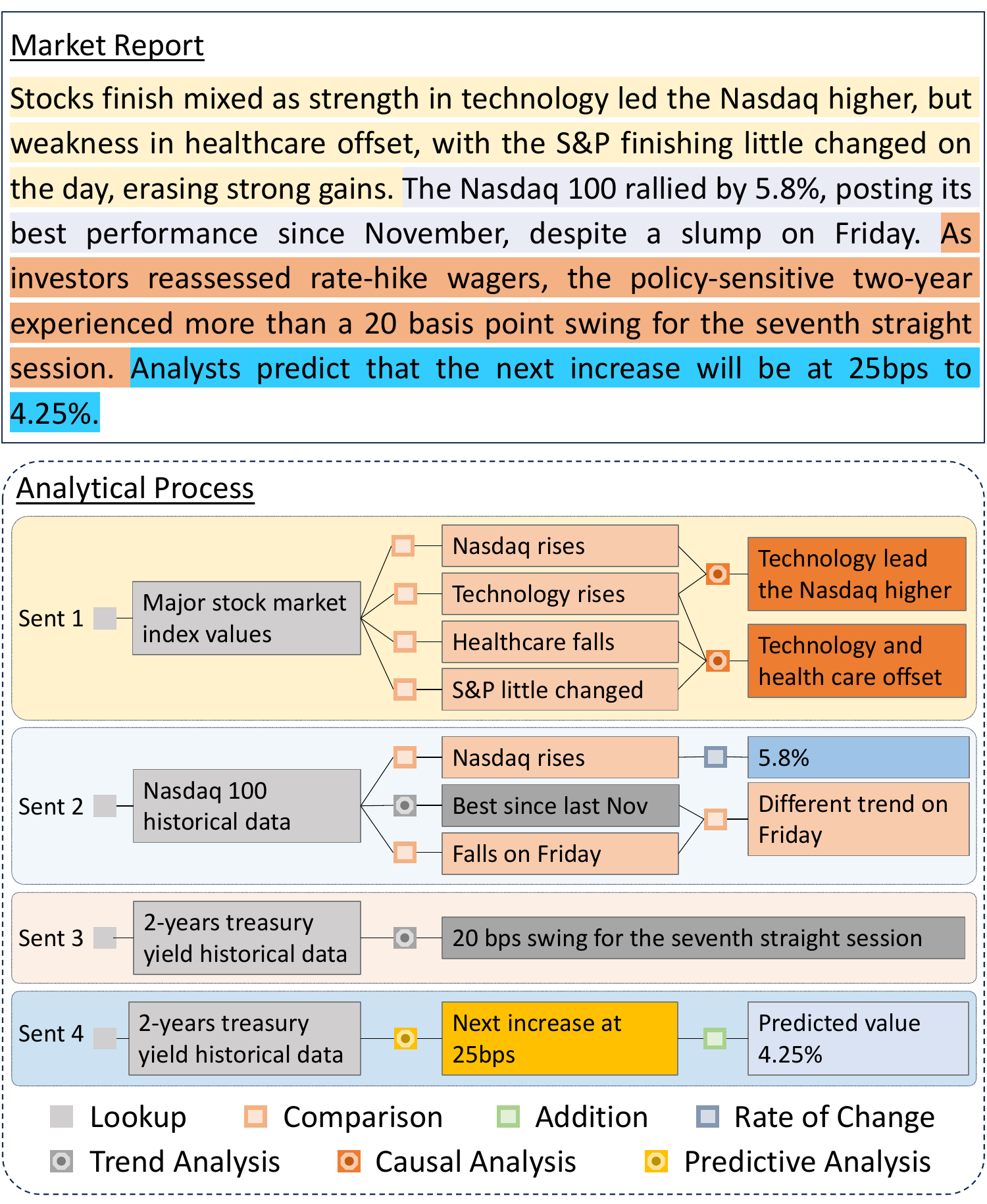}
    \caption{Example of analytical operations involved in the market report.}
    \label{fig:reasoning_graph}
    \vspace{-0.5cm}  
\end{figure}

The processed market reports of \dataset{} are narrated with analytical operations.
We identify seven most common operations, ranging from simple \textit{lookup} and basic quantitative ones such as \textit{subtraction} to more advanced analysis like \textit{causal analysis} and \textit{predictive analysis} (Table~\ref{tab:reasoning-operations}). Each category constitutes a significant portion of the report content while a sentence may involve multiple analysis, as indicated by the provided percentages. 
Figure~\ref{fig:reasoning_graph} illustrates how these operations are involved in the market reports.



\paragraph{Simple Lookup (83\%).} 
\textit{Lookup} operations involve the retrieval of data points.  They are the most common operation and serve as a prerequisite for more complex tasks. For example, \textit{trend analysis} requires lookup of market prices to identify market movement.

\paragraph{Basic Quantitative Operations (94\%).} 
Common quantitative operations include \textit{comparison}, \textit{subtraction}, and \textit{rate of change}. \textit{Comparison}, such as comparing a market index's performance to a benchmark or historical average, provide insights into relative performance and trends.
\textit{Subtraction} and \textit{rate of change} require numerical computation to obtain exact operation result. 
While \textit{subtraction} is one-hop operation, \textit{rate of change} operation is multi-hop atomic operation, posing higher numerical computation requirements.

\paragraph{Advanced Analytical Operations (84\%).} 
Advanced analytical reasoning, including  \textit{trend analysis}, \textit{causal analysis}, and \textit{predictive analysis}, forms the majority of sentences. 
These operations often require cross-referencing facts to draw conclusions, illustrating the reasoning complexity. For example, trend identification might involve analyzing moving averages or comparing past highs/lows, necessitating a system that integrates domain knowledge and performs sophisticated analytical tasks. The high prevalence of such operations highlight the importance of \dataset{}. \\

On average, we observed 2.6 operations per sentence, emphasizing the need for high-level data analytical capabilities in constructing insightful narratives. These analytical operations are not performed in isolation, but applied to specific entities and along a temporal dimension to extract meaningful insights.


\subsection{Contextual Analysis}
\label{sec:contextual-analysis}
To further understand the context of these analytical operations, we examine the entities and temporal expressions in the \dataset{} reports. This contextual analysis reveals the key focus areas and time frames shaping the reports' narrative structure.


\begin{table*}[]
\small
\resizebox{\textwidth}{!}{%
\begin{tabular}{@{}c|ccccccccccc|c@{}}
\toprule
Market & Equity & Gold & Oil & Treasury & Currency & Cattle & Corn & Dairy & Lean hog & Soybean & Wheat & Overall \\ \midrule
Data & 22 & 3 & 3 & 7 & 7 & 7 & 3 & 6 & 4 & 9 & 6 & 8.70 \\
Content & 13 & 2 & 2 & 5 & 2.5 & 7 & 3 & 5.3 & 3.7 & 5.7 & 6 & 5.69 \\
D. to C. & 0.59 & 0.67 & 0.67 & 0.71 & 0.36 & 1.00 & 1.00 & 0.88 & 0.93 & 0.63 & 1.00 & 0.65 \\ \bottomrule
\end{tabular}%
}
\caption{Unique entity counts in tabular data (Data) and the average counts in the corresponding market report content (Content) for each market subset. D. to C. (Data-to-Content) ratio measures the average percentage of the entity in the tabular data described in reports.}
\label{tab:entity_counts}
\end{table*}

\paragraph{Entities.} 
Entities in a market report form the basis for comprehensive analytical approaches, including cross-entity comparison and causal analysis (Table~\ref{tab:entity_counts}). 
Replicating the curation aspect in generating reports is important; although our reports cover many entities (8.7 on average) with high variance (3 for oil to 22 for equities), only a subset of (5.69 on average) is discussed in detail.

\paragraph{Time.}

\begin{figure}[tb]
    \centering
    \includegraphics[width=7.5cm]{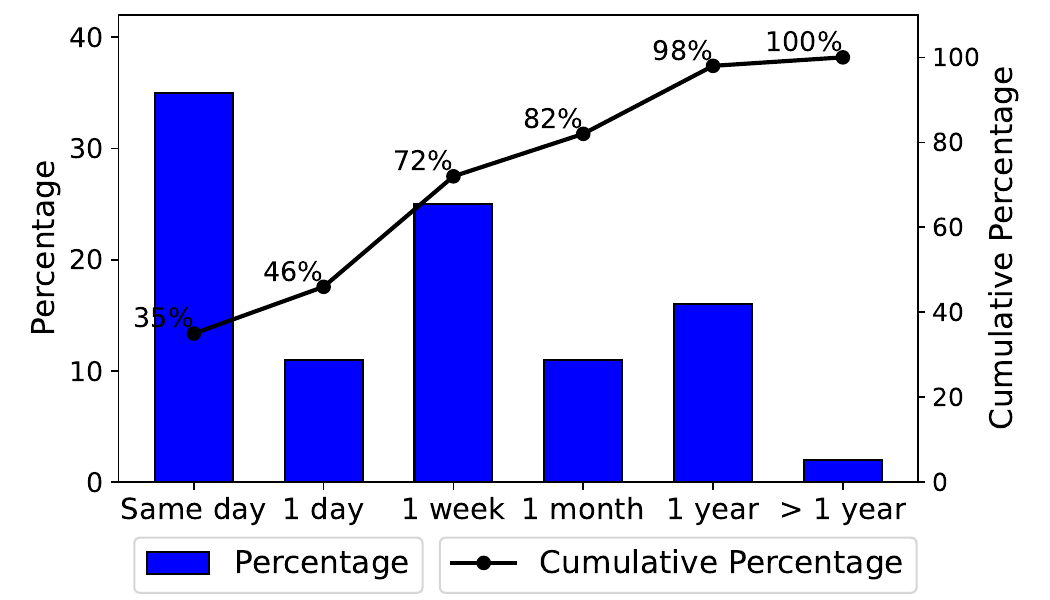}
    \caption{Distribution histogram of the time gap from the date of the referencing data to the report date. The x-axis shows the time gap, while the y-axis shows the percentage of the time gap in the (l) tabular data referencing instances, and (r) their cumulative percentage. 
    }
    \label{fig:time-difference}
     \vspace{-0.5cm}  
\end{figure}

The temporal aspect of market data is crucial in unveiling trends and projecting future movements. Like the selective detailing of entities, data spanning from the immediate day to several years is analyzed, pinpointing insightful patterns for inclusion in the  report (Figure~\ref{fig:time-difference}). The insightfulness evaluation of  model generations underscores the importance of extended tabular data, which enriches the analysis by providing a comprehensive historical context (see Section~\ref{sec:results-insightfulness}).\\


Scaling data across entities and time challenges data narration models in effectively integrating large input volumes. To navigate this complexity and convey meaningful insights, market reports must employ a domain-specific financial lexicon.

\subsection{Lexical Analysis} 
The precise and professional language in financial market reports is crucial for accurately describing entities, trends, and analytical results. This specialized vocabulary enables clear communication of complex insights derived from extensive data analysis, effectively conveying the intended message to the target audience.

\paragraph{Entity.} 
Market reports refer to entities with different terminologies (``greenback'' for US dollar) or their characteristics (``short-term bond'' for 1-year bond), highlighting the domain knowledge and linguistic versatility required of data narration.  

\paragraph{Analytical Operations.} 
Analytical results are conveyed with precision in the reports, using specific verbs like ``correct'' and ``reclaim'' to contrast current movements  against past trends, and ``pressure'' and ``push'' to indicate both direction and causality between market events. This necessitates a deep understanding of the analytical results and a strong linguistic selection capability to produce reports of comparable proficiency.

\section{Experimental Setup}


We define the task of financial data narration as follows: given market movement data $\{T_{i,j} \vert i \leq E_T, j \leq D_T\}$ with $E_T$ financial entities and $D_T$ days, where $T_{i,j}$ is the row of entity $i$ on date $j$, a \textit{data narration} model $M$ generates a report $y$ narrating the market data:
$$y = M({T_{i,j} \vert i \leq E_T, j \leq D_T})$$

\paragraph{Models.}
We explore both open-access models, specifically Llama2-7B-Chat and Llama2-13B-Chat~\citep{llama_2_paper}, alongside close-access models like GPT-3.5-Turbo and GPT-4~\citep{openai2023gpt4}. These models are renowned for their robust capabilities, particularly in terms of zero-shot generalization on new tasks.\footnote{Due to resource constraints, our main experiments focused on these selected models. To assess the generalizability of our findings, we conducted additional limited testing with other state-of-the-art models. Specifically, we ran experiments using 5 data samples each on Claude 3 Opus and Claude 3.5 Sonnet. The insights from these sample outputs were consistent with our main results discussed in Section~\ref{sec: result-discussion}, suggesting broader applicability of our findings. An example output from these additional tests is provided in the Appendix~\ref{sec:sample_generation} for reference.}

\paragraph{Evaluation Setup.} 
We evaluate real-world scenarios with no training data and examine the model's learning ability from examples. We assess both \textit{zero-shot} and \textit{fine-tuned} scenarios, splitting data based on time (first 80\% for training, remaining 20\% for validation and testing). We fine-tune with AdamW and a linear scheduler (learning rate: 1e-4, batch size: 16). We load models in 8-bit mode and fine-tune for 5 epochs on \dataset{} using LoRA due to resource constraints, and perform greedy decoding in inference.

Evaluations are conducted in two tabular data settings: (1) same-day data and (2) historical data spanning one week, to discern the influence of historical data on performance. Tabular data is linearized row-by-row for model input.


Although few-shot models have shown better performance~\citep{Wang_Yao_Kwok_Ni_2020}, input limitations of models like Llama-2 posed challenges in incorporating complete data--report pairs for in-context learning without compromising content. Future studies should consider integrating few-shot learning by examining models with extended context windows or refining methods for condensing data.

\paragraph{Metrics.}

\begin{figure}[tb]
    \centering
    \includegraphics[width=7.5cm]{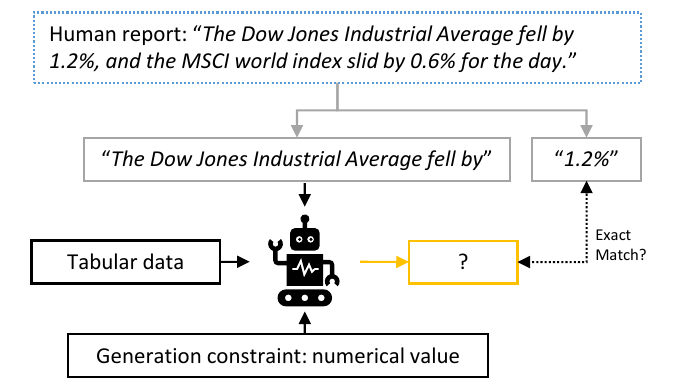}
    \caption{
    Illustration of the factuality evaluation process. We provide the model with the prefix of human reports and assess whether its predicted numerical values align with those provided by experts, thus evaluating the accuracy of the content.}
    \label{fig:accuracy_evaluation}
     \vspace{-0.5cm}  
\end{figure}

We evaluate generated text quality based on factual accuracy, insightfulness and  language style. 
(1) \textit{Factuality} is evaluated using a method inspired by MCQA tasks~\citep{DBLP:journals/corr/abs-1803-05457, Hendrycks_Burns_Basart_Zou_Mazeika_Song_Steinhardt_2021}. We first use a Named Entity Recognition model to identify numerical values in the generated text. As shown in Figure~\ref{fig:accuracy_evaluation}, the model then predicts numeric tokens given contextual information, and these predictions are compared with the original report to assess accuracy. A detailed explanation of this process is provided in Appendix~\ref{sec:accuracy_evaluation_details}.
(2) \textit{Insightfulness} is evaluated  through human assessments of freely generated model narrations. Two finance-background evaluators score them based on impact (breadth of the claim) and significance (magnitude of changes described) on a 1-5 scale, inspired by \cite{ding_quickinsights_2019}, based on sentence samples that were generated on the same tabular data. The insightfulness score is the average of the two. (3) \textit{Style} is assessed using BLEU~\citep{papineni-etal-2002-bleu} to compare N-grams with human-provided narratives. 




\section{Experimental Result and Discussion}
\label{sec: result-discussion}


Table~\ref{tab:main_results} benchmarks model performance in \textit{zero-shot} and \textit{fine-tuning} settings
under two scenarios: using same-day or one-week prior tabular data.

\begin{table*}[tb]
\centering
\resizebox{\textwidth}{!}{%
\begin{tabular}{ccccccccc}
\toprule
\textbf{Model} & \textbf{Data} & \textbf{Setting} & \textbf{Avg. Length} & \textbf{Factuality} & \textbf{Style} & \multicolumn{3}{c}{\textbf{Insightfulness}} \\
\cmidrule(lr){5-5} \cmidrule(lr){6-6} \cmidrule(lr){7-9} 
 & &  & & Acc. (\%) & BLEU (\%) & Impact & Significance & Avg. \\
\midrule
GPT-3.5-Turbo & Same day & Zero-shot & 320 & 14.58 & 3.42 & 3.26 & \textbf{2.71} & \textbf{2.98} \\
GPT-4 & Same day & Zero-shot & 423 & 25.22 & 1.96 & 3.29 & 2.51 & 2.90 \\
LlaMa2-7B & Same day & Zero-shot & 693 & 18.76 & 2.26 & 2.79 & 2.05 & 2.42 \\
LlaMa2-7B & Same day & Fine-tuned & 180 & 22.10 & 11.19 & \textbf{3.42} & 2.38 & 2.90 \\
LlaMa2-13B & Same day & Zero-shot & 502 & 20.73 & 3.40 & 3.25 & 2.52 & 2.89 \\
LlaMa2-13B & Same day & Fine-tuned & 139 & \textbf{28.93} & \textbf{14.13} & 3.40 & 2.54 & {2.97} \\
\midrule
GPT-3.5-Turbo & 1 Week & Zero-shot & 342 & 14.00 & 3.32 & \textbf{3.38} & 2.80 & {3.09} \\
GPT-4 & 1 Week & Zero-shot & 421 & \textbf{28.68} & 2.04 & 3.06 & 2.40 & 2.73 \\
LlaMa2-7B & 1 Week & Zero-shot & 405 & 11.15 & 3.34 & 2.99 & 2.48 & 2.74 \\
LlaMa2-7B & 1 Week & Fine-tuned & 136 & 11.64 & 10.47 & 3.28 & 2.39 & 2.84 \\
LlaMa2-13B & 1 Week & Zero-shot & 370 & 7.11 & 4.11 & 3.36 & \textbf{2.85} & \textbf{3.11} \\
LlaMa2-13B & 1 Week & Fine-tuned & 136 & 12.30 & \textbf{10.66} & {3.37} & 2.55 & 2.96 \\
\bottomrule
\end{tabular}%
}
\caption{Evaluation results of LLMs on \dataset{} with different settings. The \textit{Data.} specifies the time span of the tabular data provided.
The Insightfulness includes \textit{Impact} and \textit{Significance} and their average scores evaluated by human on a sample of 20 date-market combinations, with 5 sentences from each corresponding report. 
The bold text indicates the best performance among all the models in each tabular data setting. }
\label{tab:main_results}
\end{table*}

\subsection{Factuality Analysis} 

Factuality is critical for generating useful reports. 
However, results show that tested LLMs do unacceptably poorly at predicting key numbers in data narration, even with fine-tuning (sub 30\%).

\begin{figure}[tb]
    \centering
    \includegraphics[width=1\columnwidth]{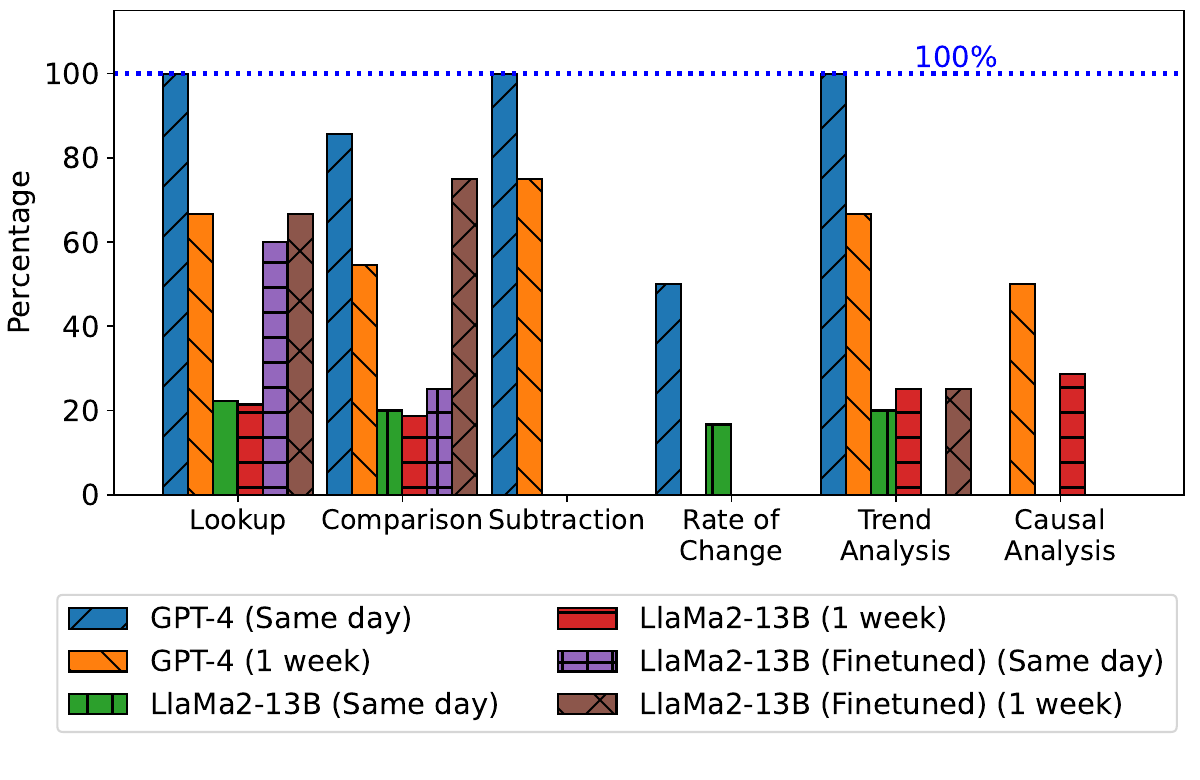}
    \caption{The accuracy of the operations in the sampled sentences generated under different settings. The green dotted line represent perfect reference, which reveals the gap between current model generations and proficient output. Predictive analysis is not included due to its unverifiable nature.}
    \label{fig:reasoning-accuracy}
     \vspace{-0.5cm}  
\end{figure}

Surprisingly, including one week of historical data detracts from performance, despite sentences describing weekly dynamics in our dataset (Figure~\ref{fig:time-difference}), possibly due to the difficulty in finding the correct value from a larger dataset, as indicated by the zero-shot \textit{Lookup} operation results in Figure~\ref{fig:reasoning-accuracy}, which impacts all analyses building on them.



It's important to note that our automated factuality evaluation method, while resembling a continuation task, was chosen due to the lack of reliable automatic evaluation techniques for freely generated narrations with extensive reasoning. To address this limitation and study the causes of low accuracy, we supplemented our automated evaluation with manual analysis. 

Specifically, we manually identified analytical operations within sampled sentences from freely generated reports and evaluated their accuracy. This analysis revealed when inaccuracies occur most frequently (Figure~\ref{fig:reasoning-accuracy}). LLMs failed to achieve required accuracy for all operations, with the error rate rising with operation complexity.

\paragraph{Causal Analysis.} 
Causal analysis accuracy depends on the complexity of information sources and the directness of the causal relationships. Simple analyses, like attributing market trends to sentiment or sector performance to index movements, rely on readily available data, making causality clearer. However, complex analyses such as linking market movements to external news, are challenging. 
LLMs occasionally succeed in correlating events like corn prices with freeze damage in the exporting country, but the overall accuracy is low.
We surmise this is due to 1) the lack of input data, forcing the reliance on potentially outdated pretraining knowledge;
2) difficulty filtering vast amounts of news to pinpoint relevant causes; and 3) the need for multi-hop processes
identifying relevant causes from vast potential arrays of causes is most challenging, requiring deep understanding of the news and their potential market impact.

\paragraph{Trend Analysis.} 
Trend analysis accuracy is tied to the duration covered. Short-term analyses 
are more accurate (52\%) due to readily available data.
Mid- and long-term analyses have lower accuracies (17.6\% and 0\%, respectively) because of unavailable extended-period data and complex longer-term analysis (e.g., the computation of 200-days moving average and Relative Strength Index).
Enhancing accuracy requires methods to access and incorporate broader historical data and improved model capabilities for accurate complex data analysis, suggesting the potential of LLMs with large token limits and insight recommendation methods.
    
\paragraph{Predictive Analysis.} 
Predictive analysis builds on causal  and trend analysis to generate deep analytical capabilities, such as forecasting gold prices to remain low (trend) due to a strengthening dollar (causality). Yet, we are the first to cover predictive analysis in data narration task to the best of our knowledge. 
However, despite its importance, the subjective nature of these predictions, unverifiable at report generation, complicates accuracy assessment. Given that even human experts err in market predictions,we believe
the logic underpinning forecasts is more crucial than its precision.\\



Our error analysis reveals the potential of \dataset{} as a comprehensive benchmark for evaluating analysis operations with different complexity in data narration. The dataset's inclusion of extended historical data aligns with real-world scenarios, challenging models to perform complex operations and advanced analyses crucial for deep insights. While our current factuality evaluation method effectively assesses the models' ability to generate faithful continuations, it may not completely disentangle factual and stylistic choices. However, this approach aligns well with the goal of faithful data narration.
Alternative evaluation methods, such as multiple-choice prediction, can be significantly impacted by the quality and order of the choices, potentially faltering in assessing LLMs' capabilities~\citep{Wang_Zhao_Qiang_Qin_Liu_2024}.


\subsection{Insightfulness Analysis}
\label{sec:results-insightfulness}

The insightfulness of market reports largely determines their value. Table~\ref{tab:main_results} shows that in zero-shot settings, larger GPT models, like GPT-4, have higher accuracy but lower insightfulness scores, while larger Llama2 models exhibit the opposite, suggesting GPT-4's fact-grounding limits in-depth analysis despite larger models' better reasoning. Fine-tuning increases impact scores, but significance scores improve only with same-day data. Reports using one week's history yield higher significance scores, proving the value of longer historical data in enhancing insight depth.

To understand the possible contribution of the analytical operations, we match the scores with the operations identify during content accuracy evaluation. Table~\ref{tab:reasoning-results} presents average scores for sentences involved various analysis operations.

\begin{table}[tb]
\centering
\resizebox{6cm}{!}{%
\begin{tabular}{@{}cccc@{}}
\toprule
\textbf{Operation} & \textbf{Impact} & \textbf{Significance} & \textbf{Avg.} \\ \midrule
Causal Analysis & 3.61 & 3.55 & 3.58 \\
Predictive Analysis & 3.50 & 3.50 & 3.50 \\
Trend Analysis & 3.55 & 3.18 & 3.37 \\
Rate of Change & 3.58 & 2.96 & 3.27 \\
Comparison & 3.52 & 2.78 & 3.15 \\
Lookup & 3.40 & 2.53 & 2.97 \\
Subtraction & 3.18 & 2.22 & 2.70 \\
Others & 2.00 & 1.50 & 1.75 \\ \bottomrule
\end{tabular}%
}
\caption{Average impact and significance scores of analysis sentences, ordered by overall insightfulness. }
\label{tab:reasoning-results}
\end{table}

\paragraph{Impact.} 
The impact score of a sentence in a financial report depends on the scope and relevance of the information. Sentences describing broader market trends or analyzing groups of entities receive higher scores, providing a more comprehensive view. For example, sentences discussing "corn futures" as a group are typically scored as 4, while those mentioning specific entities like "corn Apr future" are scored as 3. Complex analyses building upon low-level operations, which focus on specific entity, yield higher average scores by combining information from multiple entities or time periods to reveal overarching patterns and insights.

\paragraph{Significance.}
Sentences with advanced analytical operations are rated highly significant because significant market changes often require in-depth examinations, such as trend or causal analysis. These methods provide insights beyond immediate market movements, like predictive analysis offering actionable information for investment decisions, making them crucial for comprehensive market analysis. The gap between impact and significance scores for trend analysis is due to the large portion of short-term trends described in GPT-4 generation, yielding higher accuracy but less insight.\\


The demonstrated importance of advanced analytical operations in generating insightful content underscores the value of \dataset{} as a benchmark for data narration. 

We also perform model-based evaluation using win rates over human-generated reports as judged by GPT-4. 
However, a notable gap emerges between model-based evaluations and human assessments. Instead, it is found to strongly correlate with report length with based on the 12 sets of experiment results, 
suggesting a model bias favoring longer reports (See Appendix~\ref{sec:insight_winrate}). 


\subsection{Style Analysis}

\begin{table}[]
\resizebox{7.5cm}{!}{%
\begin{tabular}{@{}ccccc@{}}
\toprule
\textbf{Model} & \textbf{Setting} & \textbf{BLEU} & \textbf{Verb} & \textbf{Entity} \\ \midrule
LlaMa2-7B & Zero-shot & 3.34 & 0.565 & 0.78 \\
LlaMa2-7B & Fine-tuned & 10.47 & 0.729 & 0.859 \\
LlaMa2-13B & Zero-shot & 4.11 & 0.535 & 0.715 \\
LlaMa2-13B & Fine-tuned & \textbf{10.66} & \textbf{0.741} & \textbf{0.872} \\
GPT-3.5-Turbo & Zero-shot & 3.32 & 0.539 & 0.798 \\
GPT-4 & Zero-shot & 2.04 & 0.512 & 0.524 \\ \bottomrule
\end{tabular}%
}
\caption{The BLEU scores, and cosine similarities of verb and entity contained comparing model generated reports with the human created ones. The results here are for generations with 1 week historical data setting. We omit the result for same day setting as they show similar pattern.}
\label{tab:lexical_analysis}
\end{table}

Style analysis is crucial for evaluating models' performance on \dataset{}, as it assesses their ability to generate market reports resembling human experts' writing style, reflecting their capacity to produce informative, readable, and domain-consistent reports.

Fine-tuned LLaMa2 models significantly improve in capturing the desired writing style, with a threefold BLEU score increase over the base model (Table~\ref{tab:main_results}), suggesting \dataset{}' effectiveness in guiding lexical choices. GPT-4 exhibits the lowest BLEU score, highlighting \dataset{}' unique challenges compared to general NLP benchmarks. The cosine similarity~\citep{Pradhan_Gyanchandani_Wadhvani_2015} of entities and verbs used by the models (Table~\ref{tab:lexical_analysis}) further supports these findings, with the fine-tuned LLaMa2 achieving higher similarity scores with human experts. These findings underscore the importance of domain-specific fine-tuning for generating high-quality market reports.

This style analysis relies on automated evaluation metrics, such as BLEU scores and cosine similarity, which provide valuable insights into models' performance. Although these metrics may not capture all aspects of the generated reports, they offer a scalable and efficient way to assess models' ability to generate market reports of the desired style. Future work could benefit from incorporating human evaluation to provide qualitative and nuanced feedback, complementing these automated metrics.

\section{Conclusion \& Future Work}
\label{sec:conclusion}

We identify a crucial gap in financial data narration: the lack of benchmarks that offer deep insights, essential for real-world applications. To address this, we introduce \dataset{}, a novel dataset  covers complex analytical operations, and enhanced by extensive data for more impactful and significant insights and domain-specific languages for higher proficiency. 
The complexity of \dataset{} poses significant challenges to state-of-the-art models, in terms of both their poor accuracy and the lack of insight on their generated text. 
This complexity arises from the requirements of performing complex analytical operations, incorporating large input data, and accessing relevant knowledge. 

Moving forward, we propose three focused areas to advance financial data narration. Firstly, refining the analysis by using methods like DataShot~\citep{Wang_Sun_Zhang_Cui_Xu_Ma_Zhang_2020} and
Table2analysis~\citep{zhou_table2analysis_2020} to recommend insights with specific metrics as an intermediate step prior text narrative generation to enhance analytical quality. Secondely, integrating visuals into narratives, with methods like Foresight~\citep{DBLP:journals/corr/DemiralpHPP17}, 
Voder~\citep{Srinivasan_Drucker_Endert_Stasko_2019}, and
Calliope~\citep{Shi_Xu_Sun_Shi_Cao_2021}, to improve storytelling. Lastly, developing new automated evaluation focused on accuracy and insight quality using table fact-checking models~\citep{Li_He_table_gpt_2023, Müller_tapas_2021, Gu_Fan_pasta_2022} and insight evaluation frameworks~\citep{ding_quickinsights_2019, zhou_table2analysis_2020} respectively. 

\section*{Ethics Statement}
We have thoroughly investigated the legal aspects of using the scraped data and are confident that our dataset can be released without infringing copyright laws. The publishers' terms of use allow non-commercial use, and robots.txt files permit web scraping. For tabular data, we only release extraction scripts to ensure compliance with copyright regulations. We have taken utmost care to respect the intellectual property rights of the original data providers while creating a valuable resource for the research community.

Our study involved voluntary participation from former colleagues without financial compensation. We designed evaluation tasks to align with participants' professional expertise and implemented data anonymization to ensure privacy and confidentiality.

We acknowledge our technology's potential impact on financial analytics and emphasize responsible use. Key considerations include:

\begin{itemize}
    \item Employment effects: As AI-generated reports advance, we must address potential impacts on financial sector jobs and promote human-AI collaboration.
    \item Human oversight: We advocate for maintaining human expertise alongside AI-generated reports. Professionals should review and validate AI outputs for accuracy and context.
    \item Transparency: We recommend clearly disclosing the use of AI-generated content in financial communications to maintain trust and inform stakeholders.
\end{itemize}

We are committed to responsible AI development in financial analytics and encourage users of our dataset and technology to implement appropriate ethical safeguards.

\section*{Limitations}
Firstly, the method and dataset are primarily designed for languages with limited morphology, such as English. Secondly, our \dataset{} dataset is specifically focused on market movement data, which represents only 52\% of the content for a human generated report. Further research can explore the inclusion of market context and external news, to provide more in-depth analysis especially for causal analysis and predictive analysis.  Lastly, our \dataset{} dataset focuses on textual narratives, while charts are found to be useful for a market report in real-world. It would be beneficial for future studies to aim for a multi-modal report to provide a more useful reports.


\bibliography{custom}
\clearpage

\appendix
\onecolumn
\section{Sentence Classification}
\label{sec:appendix-sentence_classification}

\subsection{Sentence Classes}
We classify the raw market reports into the following categories based on their main source of information.

\begin{itemize}
    \item \textit{Market Movements}: describing tangible shifts such as asset prices or trends;
    \item \textit{Market Context}: providing broader understanding and context of the market dynamics;
    \item \textit{External Events and Influence}: highlighting outside events that impact the market;
    \item \textit{Prediction and Suggestion}: encompassing forward-looking statements based on current data and analysis.
\end{itemize}

The classification was done using ChatGPT with in-context learning:

\subsection{Classification prompt}
\lstset{
    style=mystyle,
    basicstyle=\ttfamily\scriptsize,
    backgroundcolor=\color{white},
    stringstyle=\color{black},
    keywordstyle=\color{black},
    breaklines=false,
    keepspaces=false
}
\begin{lstlisting}[language=Python]
Predict the information type of given sentences based on the textual context provided in the passage. The 
possible information types are market movements, market context, external events and influence, and 
prediction and suggestion. The definitions and examples of the information types are listed below.
1. Market Movements: the changes in the prices, values, or trends of financial assets, such as stocks, 
commodities, or indices. Example: price fluctuations, percentage changes, or historical comparisons
2. Market Context: the broader factors, sentiment, or conditions that impact financial markets or 
assets. Example: market sentiment, or investor behavior
3. External Events and Influence: developments or occurrences outside the market itself that have a direct 
or indirect impact on financial assets or market conditions. Example: news about economic indicators, 
geopolitical events, central bank policies, regulatory changes, or corporate announcements
4. Prediction and Suggestions: projection or speculation about future market movements, trends, or 
events based on current data, analysis, or expert opinions. Example: trend forecasts, or action 
recommendations

(*@\color{codepurple}{\textbf{Passage}}@*): 
Cattle prices saw price weakness to start the week, as both live and feeder cattle post marginal losses. 
Feb cattle lost 0.300 to 137.675, and Apr cattle was 0.275 lower to 141.850. In the feeder market, Mar 
feeders dropped 0.950 to 165.425. The cattle market stays in consolidation trade on top of key support 
levels, working a narrow range. Trendline support held Apr live cattle around the $140 level, and
resistance at $142 kept prices in check again this afternoon. Apr cattle have traded within this range for
9 consecutive trading sessions. Prices are looking for direction, but with the near-term trend working 
lower, a possible break to the downside is still a possibility. The recent spike in COVID cases has 
pressured the cattle market on its impacts on cattle numbers and supply chain, but the market is more 
optimistic that this will be a short-term issue. Estimated slaughter today may be showing some movement 
out of that concern. For today, 117,000 head was forecasted for kill, up 3,000 head from last week. Cash 
trade is typically slow to start the week, but some very light trade did occur at $137, but not enough to 
establish a clear trend. More trade will develop later on in the week. At midday, choice carcasses added 
1.43 to 289.29 and select was 0.78 firmer to 277.83. Load count was light at 77 loads as the trend higher 
in retail beef continues. The weakness in live cattle, and mixed grain trade kept the feeder market 
pressured on the day. Jan feeders expire on January 27 and are closely tied to the cash index. The Feeder 
Index was 0.21 lower to 161.80. The cattle market is still trending higher overall, but near-term prices 
are in a consolidation pattern, looking for a reason to move either higher or lower.
(*@\color{codepurple}{\textbf{Sentence}}@*): 
The cattle market stays in consolidation trade on top of key support levels, working a narrow range.
(*@\color{codepurple}{\textbf{Information Type}}@*): 
market movements


(*@\color{codepurple}{\textbf{Passage}}@*): 
Asia-Pacific stocks largely rose on Thursday. China's consumer price index and producer price index for
September were released on Thursday. Singapore's Straits Times index gained 0.24% as of 3:27 p.m. local
time, recovering from earlier losses after the country's central bank unexpectedly tightened monetary
policy on Thursday.
(*@\color{codepurple}{\textbf{Sentence}}@*): 
Singapore's Straits Times index gained 0.24% as of 3:27 p.m. local time, recovering from earlier losses
after the country's central bank unexpectedly tightened monetary policy on Thursday.
(*@\color{codepurple}{\textbf{Information Type}}@*): 
market movements


(*@\color{codepurple}{\textbf{Passage}}@*): 
U.S. equity markets bounced back strongly on Thursday as upbeat economic data and stellar corporate 
earnings results boosted market sentiment. The S&P 500 gained 1% to close at another record high of 
4596.42. The Nasdaq gained 1.4% to close at 15448.12, and the Dow gained 0.7% to close at 35730.48. Since 
earnings season began, 82% of the companies that make up the S&P 500 has been able to report earnings that 
beat analyst estimates. The U.S. GDP grew by 2% ,quarter over quarter, marking the weakest quarter of 
growth since mid-2020. A surge in COVID cases and the supply chain crunch both hindered the growth over 
the past quarter. On the other hand, initial jobless claims figure hit a fresh pandemic low at 281,000. 
The 10 year treasury yield increased slightly to settle at 1.578% and the 30 year treasury yield increased 
slightly as well to settle at 1.979%. Facebook's CEO, Mark Zuckerberg, has announced that, beginning on
December 1st, Facebook will be rebranded as Meta Platforms Inc and will be switching the ticker FB to 
MVRS. The new parent company will be devoted to creating a more immersive experience of the world wide web 
by combining virtual reality and building a virtual world where all media sources can be combined and 
utilized.
(*@\color{codepurple}{\textbf{Sentence}}@*): 
Since earnings season began, 82% of the companies that make up the S&P 500 has been able to report 
earnings that beat analyst estimates.
(*@\color{codepurple}{\textbf{Information Type}}@*): 
market context


(*@\color{codepurple}{\textbf{Passage}}@*): 
The Pound continues to fall, threatening to drop below $1.30 during early Wednesday trading, as hopes seem 
slim that an agreement to deliver an orderly Brexit will be reached by Prime Minister Theresa May and 
opposition leader Jeremey Corbyn. Both politicians have been under pressure to deliver a solution, after 
suffering a humiliating defeat at the recent local British election, which was interpreted as a reaction 
to the disappointment generated by the Brexit related shenanigans at Westminster. For a while there were 
hopes that talks would lead to a resolution and that supported Sterling. However, the latest developments 
have been met with dismay by the markets, triggering a drop for the Pound, as it becomes increasingly 
clear that we are in for more of the same and no agreement is likely to be reached anytime soon.
(*@\color{codepurple}{\textbf{Sentence}}@*): 
Both politicians have been under pressure to deliver a solution, after suffering a humiliating defeat at 
the recent local British election, which was interpreted as a reaction to the disappointment generated by 
the Brexit related shenanigans at Westminster.
(*@\color{codepurple}{\textbf{Information Type}}@*): 
external events and influence


(*@\color{codepurple}{\textbf{Passage}}@*): 
The short-term outlook for gold appears mixed, with prices suffering from the risk-on mood now prevailing
in financial markets, which pushed investors to bet on riskier asset. Traders are now waiting for Mario 
Draghi's speech, to understand if he will be able to fulfil the expectations that he raised of even lower
interest rates and more economic stimulus, which could be in the form of assets purchase. Since 
expectations for the meeting of tomorrow are quite high, the risk of a disappointment is high too, if 
Draghi does not manage to get approval from his team to go ahead with this measure in his final ECB 
meeting. The decline of gold seen in the last few days is also reflecting these expectations. In any case, 
the correction remains moderate, even if prices reached a new 1-month low. Gold is still within the values 
hit on August 13 (low 1,477, high 1,533). A break up or down of these levels could offer a new directional 
impulse, even if the main trend still appears positive.
(*@\color{codepurple}{\textbf{Sentence}}@*): 
A break up or down of these levels could offer a new directional impulse, even if the main trend still 
appears positive.
(*@\color{codepurple}{\textbf{Information Type}}@*): 
prediction and suggestion

(*@\color{codepurple}{\textbf{Passage}}@*): (*@{\color{darkblue}{\textsc{<human\_created\_report\_to\_predict>}}@*)
(*@\color{codepurple}{\textbf{Sentence}}@*): (*@{\color{darkblue}{\textsc{<sentence\_from\_report\_to\_predict>}}@*)
(*@\color{codepurple}{\textbf{Information type}}@*): 
\end{lstlisting}

\subsection{Classification results}
The sentence classification result as listed in Table~\ref{tab:info-type-stats}.

\begin{table*}[h]
\centering
\resizebox{7cm}{!}{%
\begin{tabular}{c c c c}
\hline
\textbf{Information Type}     & \textbf{Percentage} & \textbf{Sent} & \textbf{Word} \\ \hline
market movements              & 50.1\%              & 4.6           & 101.7         \\ 
prediction and suggestion     & 4.3\%               & 1.4           & 32.8          \\ 
market context                & 26.4\%              & 2.9           & 76.0          \\ 
external events and influence & 19.2\%              & 2.5           & 72.4          \\ \hline
\end{tabular}%
}
\caption{The table shows the percentage and sentence/word count for each information type in the dataset.}
\label{tab:info-type-stats}
\end{table*}

\onecolumn
\section{Report Generation Results}
\label{sec:sample_generation}
We provide the report generated by human and baseline models for same market (equity market) and report date (2023-01-24), demonstrating the conclusions about factuality, insightfulness, and text style discussed in Section~\ref{sec: result-discussion}.
\subsection{Report Generation Prompt}
\label{sec:model_generation_prompt}
\begin{lstlisting}[language=Python]
Please act as an expert financial market analyst. Please generate market report for <market> market on 
<target_report_date> that:
1. based only on the historical market data provided.
2. following the market report example provided.

(*@\color{codepurple}{\textbf{Report Date}}@*): (*@{\color{darkblue}{\textsc{<target\_report\_date>}}@*)

(*@\color{codepurple}{\textbf{Market Data}}@*): (*@{\color{darkblue}{\textsc{<tabular\_table>}}@*)

\end{lstlisting}

\subsection{Sample Input Table}

\subsubsection{Same day data}
\begin{lstlisting}
report date: 2023-01-24

market information: 
market: equity market
financial_instrument: NASDAQ Composite
historical_data: 
Financial Instrument       Date      Open      High       Low     Close Volume
    NASDAQ Composite 2023-01-24 11,304.13 11,378.15 11,284.29 11,334.27      0

market: equity market
financial_instrument: S&P 500
historical_data: 
Financial Instrument       Date     Open     High      Low    Close Volume
             S&P 500 2023-01-24 4,001.74 4,023.92 3,989.79 4,016.95    NaN

market: equity market
financial_instrument: S&P 500 Consumer Discretionary
historical_data: 
          Financial Instrument       Date    Open   High    Low  Close Volume
S&P 500 Consumer Discretionary 2023-01-24 1086.75 1096.9 1096.9 1096.9    NaN

market: equity market
financial_instrument: S&P 500 Consumer Staples
historical_data: 
    Financial Instrument       Date   Open   High    Low  Close Volume
S&P 500 Consumer Staples 2023-01-24 767.08 764.03 764.03 764.03    NaN

market: equity market
financial_instrument: S&P 500 Health Care
historical_data: 
Financial Instrument       Date    Open    High     Low   Close Volume
 S&P 500 Health Care 2023-01-24 1566.24 1556.12 1556.12 1556.12    NaN

market: equity market
financial_instrument: S&P 500 Industrials
historical_data: 
Financial Instrument       Date   Open   High    Low  Close Volume
 S&P 500 Industrials 2023-01-24 845.36 852.53 852.53 852.53    NaN

market: equity market
financial_instrument: S&P 500 Information Technology
historical_data: 
          Financial Instrument       Date    Open    High     Low   Close Volume
S&P 500 Information Technology 2023-01-24 2344.75 2343.86 2343.86 2343.86    NaN

market: equity market
financial_instrument: S&P 500 Materials
historical_data: 
Financial Instrument       Date   Open   High    Low  Close Volume
   S&P 500 Materials 2023-01-24 523.62 523.64 523.64 523.64    NaN

market: equity market
financial_instrument: S&P 500 Real Estate
historical_data: 
Financial Instrument       Date   Open   High    Low  Close Volume
 S&P 500 Real Estate 2023-01-24 248.33 248.33 248.33 248.33    NaN

market: equity market
financial_instrument: S&P 500 Communication Services
historical_data: 
          Financial Instrument       Date   Open   High    Low  Close Volume
S&P 500 Communication Services 2023-01-24 180.24 178.99 178.99 178.99    NaN

market: equity market
financial_instrument: S&P 500 Utilities
historical_data: 
Financial Instrument       Date   Open   High    Low  Close Volume
   S&P 500 Utilities 2023-01-24 352.72 353.85 353.85 353.85    NaN

market: equity market
financial_instrument: S&P 500 Financials
historical_data: 
Financial Instrument       Date   Open   High    Low  Close Volume
  S&P 500 Financials 2023-01-24 594.75 595.37 595.37 595.37    NaN

market: equity market
financial_instrument: S&P 500 Energy
historical_data: 
Financial Instrument       Date   Open   High    Low  Close Volume
      S&P 500 Energy 2023-01-24 693.91 692.41 692.41 692.41    NaN

market: equity market
financial_instrument: Dow Jones Industrial Average
historical_data: 
        Financial Instrument       Date     Open     High      Low    Close Volume
Dow Jones Industrial Average 2023-01-24 33444.72 33782.92 33310.56 33733.96    NaN

market: equity market
financial_instrument: Russell 2000
historical_data: 
Financial Instrument       Date     Open     High      Low    Close Volume
        Russell 2000 2023-01-24 1,887.81 1,892.71 1,878.33 1,885.61    NaN

market: equity market
financial_instrument: CBOE Volatility Index
historical_data: 
 Financial Instrument       Date  Open  High   Low Close Volume
CBOE Volatility Index 2023-01-24 19.89 20.47 18.91  19.2    NaN

market: equity market
financial_instrument: gold
historical_data: 
Financial Instrument       Date     Open     High      Low    Close Volume
                gold 2023-01-24 1,931.33 1,941.23 1,920.53 1,937.35   5721

market: equity market
financial_instrument: Dollar index
historical_data: 
Financial Instrument       Date    Open    High     Low   Close Volume
        Dollar index 2023-01-24 101.988 102.428 101.716 101.918      0

market: equity market
financial_instrument: 2-year treasury yield
historical_data: 
 Financial Instrument       Date Open High Low Close Volume
2-year treasury yield 2023-01-24 4.23 4.26 4.2  4.21    NaN

market: equity market
financial_instrument: 10-year treasury yield
historical_data: 
  Financial Instrument       Date Open High  Low Close Volume
10-year treasury yield 2023-01-24 3.45 3.47 3.45  3.47    NaN

\end{lstlisting}

\subsubsection{1 week data}

\begin{lstlisting}
report date: 2023-01-24

market information: 
market: equity market
financial_instrument: NASDAQ Composite
historical_data: 
Financial Instrument       Date      Open      High       Low     Close Volume
    NASDAQ Composite 2023-01-18 11,165.88 11,223.41 10,952.05 10,957.01      0
    NASDAQ Composite 2023-01-19 10,895.92 10,932.52 10,804.57 10,852.27      0
    NASDAQ Composite 2023-01-20 10,922.53 11,143.17 10,885.65 11,140.43      0
    NASDAQ Composite 2023-01-23 11,161.97 11,405.50 11,144.03 11,364.41      0
    NASDAQ Composite 2023-01-24 11,304.13 11,378.15 11,284.29 11,334.27      0

market: equity market
financial_instrument: S&P 500
historical_data: 
Financial Instrument       Date     Open     High      Low    Close Volume
             S&P 500 2023-01-18 4,002.25 4,014.16 3,926.59 3,928.86    NaN
             S&P 500 2023-01-19 3,911.84 3,922.94 3,885.54 3,898.85    NaN
             S&P 500 2023-01-20 3,909.04 3,972.96 3,897.86 3,972.61    NaN
             S&P 500 2023-01-23 3,978.14 4,039.31 3,971.64 4,019.81    NaN
             S&P 500 2023-01-24 4,001.74 4,023.92 3,989.79 4,016.95    NaN

market: equity market
financial_instrument: S&P 500 Consumer Discretionary
historical_data: 
          Financial Instrument       Date    Open    High     Low   Close Volume
S&P 500 Consumer Discretionary 2023-01-18    1100 1073.99 1073.99 1073.99    NaN
S&P 500 Consumer Discretionary 2023-01-19 1066.06 1055.81 1055.81 1055.81    NaN
S&P 500 Consumer Discretionary 2023-01-20 1058.12 1081.81 1081.81 1081.81    NaN
S&P 500 Consumer Discretionary 2023-01-23 1085.35 1098.79 1098.79 1098.79    NaN
S&P 500 Consumer Discretionary 2023-01-24 1086.75  1096.9  1096.9  1096.9    NaN

market: equity market
financial_instrument: S&P 500 Consumer Staples
historical_data: 
    Financial Instrument       Date   Open   High    Low  Close Volume
S&P 500 Consumer Staples 2023-01-18 778.89 760.45 760.45 760.45    NaN
S&P 500 Consumer Staples 2023-01-19 760.38 752.72 752.72 752.72    NaN
S&P 500 Consumer Staples 2023-01-20 753.24 758.85 758.85 758.85    NaN
S&P 500 Consumer Staples 2023-01-23 759.88 761.11 761.11 761.11    NaN
S&P 500 Consumer Staples 2023-01-24 767.08 764.03 764.03 764.03    NaN

market: equity market
financial_instrument: S&P 500 Health Care
historical_data: 
Financial Instrument       Date    Open    High     Low   Close Volume
 S&P 500 Health Care 2023-01-18 1571.65 1549.37 1549.37 1549.37    NaN
 S&P 500 Health Care 2023-01-19 1549.37 1552.99 1552.99 1552.99    NaN
 S&P 500 Health Care 2023-01-20 1552.99 1561.79 1561.79 1561.79    NaN
 S&P 500 Health Care 2023-01-23 1561.79 1566.24 1566.24 1566.24    NaN
 S&P 500 Health Care 2023-01-24 1566.24 1556.12 1556.12 1556.12    NaN

market: equity market
financial_instrument: S&P 500 Industrials
historical_data: 
Financial Instrument       Date   Open   High    Low  Close Volume
 S&P 500 Industrials 2023-01-18  862.2 843.87 843.87 843.87    NaN
 S&P 500 Industrials 2023-01-19 839.04 826.29 826.29 826.29    NaN
 S&P 500 Industrials 2023-01-20 828.07 837.89 837.89 837.89    NaN
 S&P 500 Industrials 2023-01-23 839.53 847.05 847.05 847.05    NaN
 S&P 500 Industrials 2023-01-24 845.36 852.53 852.53 852.53    NaN

market: equity market
financial_instrument: S&P 500 Information Technology
historical_data: 
          Financial Instrument       Date    Open    High     Low   Close Volume
S&P 500 Information Technology 2023-01-18 2286.94 2257.63 2257.63 2257.62    NaN
S&P 500 Information Technology 2023-01-19 2257.62 2231.84 2231.84 2231.84    NaN
S&P 500 Information Technology 2023-01-20 2231.84 2292.54 2292.54 2292.54    NaN
S&P 500 Information Technology 2023-01-23 2292.54 2344.75 2344.75 2344.75    NaN
S&P 500 Information Technology 2023-01-24 2344.75 2343.86 2343.86 2343.86    NaN

market: equity market
financial_instrument: S&P 500 Materials
historical_data: 
Financial Instrument       Date   Open   High    Low  Close Volume
   S&P 500 Materials 2023-01-18 526.62 515.28 515.28 515.28    NaN
   S&P 500 Materials 2023-01-19 513.35 511.31 511.31 511.31    NaN
   S&P 500 Materials 2023-01-20 513.45 521.77 521.77 521.77    NaN
   S&P 500 Materials 2023-01-23 520.28 523.43 523.43 523.43    NaN
   S&P 500 Materials 2023-01-24 523.62 523.64 523.64 523.64    NaN

market: equity market
financial_instrument: S&P 500 Real Estate
historical_data: 
Financial Instrument       Date   Open   High    Low  Close Volume
 S&P 500 Real Estate 2023-01-18 244.92 244.92 244.92 244.92    NaN
 S&P 500 Real Estate 2023-01-19 243.82 243.82 243.82 243.82    NaN
 S&P 500 Real Estate 2023-01-20 246.74 246.74 246.74 246.74    NaN
 S&P 500 Real Estate 2023-01-23 247.33 247.33 247.33 247.33    NaN
 S&P 500 Real Estate 2023-01-24 248.33 248.33 248.33 248.33    NaN

market: equity market
financial_instrument: S&P 500 Communication Services
historical_data: 
          Financial Instrument       Date   Open   High    Low  Close Volume
S&P 500 Communication Services 2023-01-18  170.4 168.82 168.82 168.82    NaN
S&P 500 Communication Services 2023-01-19 168.82 170.34 170.34 170.34    NaN
S&P 500 Communication Services 2023-01-20 170.34 177.09 177.09 177.09    NaN
S&P 500 Communication Services 2023-01-23 177.09 180.24 180.24 180.24    NaN
S&P 500 Communication Services 2023-01-24 180.24 178.99 178.99 178.99    NaN

market: equity market
financial_instrument: S&P 500 Utilities
historical_data: 
Financial Instrument       Date   Open   High    Low  Close Volume
   S&P 500 Utilities 2023-01-18 363.05 353.46 353.46 353.46    NaN
   S&P 500 Utilities 2023-01-19 353.04 349.93 349.93 349.93    NaN
   S&P 500 Utilities 2023-01-20 349.47 351.99 351.99 351.99    NaN
   S&P 500 Utilities 2023-01-23 350.68 352.14 352.14 352.14    NaN
   S&P 500 Utilities 2023-01-24 352.72 353.85 353.85 353.85    NaN

market: equity market
financial_instrument: S&P 500 Financials
historical_data: 
Financial Instrument       Date   Open   High    Low  Close Volume
  S&P 500 Financials 2023-01-18 596.59 585.64 585.64 585.64    NaN
  S&P 500 Financials 2023-01-19 585.64 578.61 578.61 578.61    NaN
  S&P 500 Financials 2023-01-20 578.62 588.16 588.16 588.16    NaN
  S&P 500 Financials 2023-01-23 588.16 594.75 594.75 594.75    NaN
  S&P 500 Financials 2023-01-24 594.75 595.37 595.37 595.37    NaN

market: equity market
financial_instrument: S&P 500 Energy
historical_data: 
Financial Instrument       Date   Open   High    Low  Close Volume
      S&P 500 Energy 2023-01-18 691.07 678.87 678.87 678.87    NaN
      S&P 500 Energy 2023-01-19 678.87 686.38 686.38 686.38    NaN
      S&P 500 Energy 2023-01-20 686.38 695.32 695.32 695.32    NaN
      S&P 500 Energy 2023-01-23 695.32 693.91 693.91 693.91    NaN
      S&P 500 Energy 2023-01-24 693.91 692.41 692.41 692.41    NaN

market: equity market
financial_instrument: Dow Jones Industrial Average
historical_data: 
        Financial Instrument       Date     Open     High      Low    Close Volume
Dow Jones Industrial Average 2023-01-18 33948.49 34016.53  33269.9 33296.96    NaN
Dow Jones Industrial Average 2023-01-19 33171.35 33227.49 32982.05 33044.56    NaN
Dow Jones Industrial Average 2023-01-20 33073.46 33381.95 32948.93 33375.49    NaN
Dow Jones Industrial Average 2023-01-23 33439.56 33782.88 33316.25 33629.56    NaN
Dow Jones Industrial Average 2023-01-24 33444.72 33782.92 33310.56 33733.96    NaN

market: equity market
financial_instrument: Russell 2000
historical_data: 
Financial Instrument       Date     Open     High      Low    Close Volume
        Russell 2000 2023-01-18 1,890.09 1,903.87 1,854.32 1,854.36    NaN
        Russell 2000 2023-01-19 1,846.35 1,846.35 1,825.58 1,836.35    NaN
        Russell 2000 2023-01-20 1,847.68 1,867.34 1,836.63 1,867.34    NaN
        Russell 2000 2023-01-23 1,868.96 1,896.20 1,867.49 1,890.77    NaN
        Russell 2000 2023-01-24 1,887.81 1,892.71 1,878.33 1,885.61    NaN

market: equity market
financial_instrument: CBOE Volatility Index
historical_data: 
 Financial Instrument       Date  Open  High   Low Close Volume
CBOE Volatility Index 2023-01-18 19.28 20.58 18.71 20.34    NaN
CBOE Volatility Index 2023-01-19 20.43 21.71 20.17 20.52    NaN
CBOE Volatility Index 2023-01-20 20.28  20.7 19.41 19.85    NaN
CBOE Volatility Index 2023-01-23 20.21 20.33 19.55 19.81    NaN
CBOE Volatility Index 2023-01-24 19.89 20.47 18.91  19.2    NaN

market: equity market
financial_instrument: gold
historical_data: 
Financial Instrument       Date     Open     High      Low    Close Volume
                gold 2023-01-18 1,908.63 1,924.11 1,896.95 1,904.18   5484
                gold 2023-01-19 1,904.17 1,934.24 1,901.37 1,932.11   5522
                gold 2023-01-20 1,932.14 1,935.63 1,922.01 1,926.47   5671
                gold 2023-01-23 1,929.99 1,934.86 1,913.52 1,931.36   5540
                gold 2023-01-24 1,931.33 1,941.23 1,920.53 1,937.35   5721

market: equity market
financial_instrument: Dollar index
historical_data: 
Financial Instrument       Date    Open    High     Low   Close Volume
        Dollar index 2023-01-18 102.464 102.899 101.528 102.363      0
        Dollar index 2023-01-19 102.383 102.481 102.016 102.058      0
        Dollar index 2023-01-20 102.076 102.552 101.938 102.012      0
        Dollar index 2023-01-23 101.936 102.275 101.589 102.138      0
        Dollar index 2023-01-24 101.988 102.428 101.716 101.918      0

market: equity market
financial_instrument: 2-year treasury yield
historical_data: 
 Financial Instrument       Date Open High  Low Close Volume
2-year treasury yield 2023-01-18  4.2 4.22 4.07  4.08    NaN
2-year treasury yield 2023-01-19 4.09 4.14 4.04  4.12    NaN
2-year treasury yield 2023-01-20 4.13  4.2 4.13  4.18    NaN
2-year treasury yield 2023-01-23 4.18 4.24 4.16  4.24    NaN
2-year treasury yield 2023-01-24 4.23 4.26  4.2  4.21    NaN

market: equity market
financial_instrument: 10-year treasury yield
historical_data: 
  Financial Instrument       Date Open High  Low Close Volume
10-year treasury yield 2023-01-18 3.37 3.38 3.32  3.32    NaN
10-year treasury yield 2023-01-19  3.4 3.43 3.39  3.41    NaN
10-year treasury yield 2023-01-20  3.4  3.5 3.39  3.48    NaN
10-year treasury yield 2023-01-22 3.49 3.49 3.46  3.47    NaN
10-year treasury yield 2023-01-23 3.52 3.52 3.51  3.52    NaN
10-year treasury yield 2023-01-24 3.45 3.47 3.45  3.47    NaN

\end{lstlisting}

\subsection{Human-created Report}
\lstset{
    style=mystyle,
    basicstyle=\ttfamily\scriptsize,
    backgroundcolor=\color{white},
    stringstyle=\color{black},
    keywordstyle=\color{black},
    breaklines=false,
    keepspaces=false
}
\begin{lstlisting}
Stocks slipped late day, but bulls remain in charge, with the S&P 500 holding above 4,000 and manage to hold 
big gains from the last two trading days into a big earnings week. Eight of the eleven S&P sectors closed 
higher (energy, healthcare, communications dipped) and breadth continues to push higher on the NYSE. That 
number likely weakened today as UNP shares fell in transports on weaker results, as did MMM in industrials 
(and lower guide), VZ mixed results and lower guide in telco, JNJ results slight beat but enough to rally 
shares. On the flip side TRV rebounded on in-line results after guiding lower last week and PCAR a standout 
in industrials on its results. The Nasdaq Composite is on track for its best start to a year\u2013measured 
by its performance over the first 15 trading days\u2013since 2001 (+8.47%). THE CBOE volatility index (VIX) 
extends weakness, dropping over 4% below 19 today (still no fear in mkt). Treasury yields slide along with 
the dollar, while energy prices dip.
\end{lstlisting}

\subsection{Sample Model-generated Reports}

\subsubsection{GPT-3.5-Turbo (same day data)}
\begin{lstlisting}
January 24, 2023 Market Report:\n\nThe equity market showed mixed results on January 24, 2023. The NASDAQ 
Composite opened at 11,304.13 and closed at 11,334.27, indicating a slight increase of 0.27%. The S&P 500 
opened at 4,001.74 and closed at 4,016.95, indicating a marginal increase of 0.38%. The Dow Jones Industrial 
Average opened at 33,444.72 and closed at 33,733.96, indicating a moderate increase of 0.87%. The Russell 
2000 opened at 1,887.81 and closed at 1,885.61, indicating a marginal decrease of 0.12%.\n\nThe CBOE 
Volatility Index opened at 19.89 and closed at 19.20, indicating a decrease of 3.47%. Gold opened at 
1,931.33 and closed at 1,937.35, indicating a marginal increase of 0.31%. The Dollar Index opened at 101.988 
and closed at 101.918, indicating a marginal decrease of 0.07%.\n\nThe 2-year Treasury yield opened at 4.23 
and closed at 4.21, indicating a marginal decrease of 0.47%. The 10-year Treasury yield opened at 3.45 and 
closed at 3.47, indicating a marginal increase of 0.58%.\n\nOverall, the equity market showed mixed results, 
with some indices showing marginal increases and others showing marginal decreases. The CBOE Volatility 
Index decreased, indicating a decrease in market volatility. The 2-year Treasury yield decreased, while the 
10-year Treasury yield increased, indicating a slight flattening of the yield curve.
\end{lstlisting}

\subsubsection{GPT-4 (same day data)}
\begin{lstlisting}
Equity Market Report - January 24, 2023\n\nThe equity market experienced mixed performance on January 24, 
2023, with some indices showing gains while others experienced losses. The NASDAQ Composite closed at 
11,334.27, a slight increase from its opening value of 11,304.13. The S&P 500 also saw a modest gain, 
closing at 4,016.95 after opening at 4,001.74.\n\nSector performance within the S&P 500 was varied. The 
Consumer Discretionary sector closed at 1,096.90, showing no change from its opening value. The Consumer 
Staples sector experienced a slight decrease, closing at 764.03 after opening at 767.08. The Health Care 
sector also saw a decline, closing at 1,556.12 after opening at 1,566.24. The Industrials sector remained 
unchanged, closing at 852.53. The Information Technology sector experienced a minor decrease, closing at 
2,343.86 after opening at 2,344.75. The Materials sector remained stable, closing at 523.64. The Real Estate 
sector also remained unchanged, closing at 248.33. The Communication Services sector saw a decrease, closing 
at 178.99 after opening at 180.24. The Utilities sector experienced a slight increase, closing at 353.85 
after opening at 352.72. The Financials sector remained stable, closing at 595.37. The Energy sector 
experienced a minor decrease, closing at 692.41 after opening at 693.91.\n\nThe Dow Jones Industrial Average 
experienced an increase, closing at 33,733.96 after opening at 33,444.72. The Russell 2000 closed at 
1,885.61, showing a slight decline from its opening value of 1,887.81.\n\nThe CBOE Volatility Index, a 
measure of market volatility, closed at 19.2 after reaching a high of 20.47 and a low of 18.91.\n\nIn the 
commodities market, gold experienced an increase, closing at 1,937.35 after opening at 1,931.33. The Dollar 
index, a measure of the U.S. dollar's value relative to a basket of foreign currencies, closed at 101.918 
after opening at 101.988.\n\nIn the bond market, the 2-year treasury yield closed at 4.21, showing a slight 
decrease from its opening value of 4.23. The 10-year treasury yield experienced an increase, closing at 3.47 
after opening at 3.45.\n\nIn summary, the equity market on January 24, 2023, experienced mixed performance, 
with some indices and sectors showing gains while others experienced losses. Market volatility remained 
relatively stable, and the bond market experienced minor fluctuations.
\end{lstlisting}

\subsubsection{Claude 3 Opus (same day data)}
\begin{lstlisting}
Here is a market report for the equity market on 2023-01-24 based on the historical data provided:\n\n# 
Equity Market Report\n## Date: 2023-01-24\n\n### Key Takeaways\n- The major US stock indices closed higher,
with the Dow Jones Industrial Average and S&P 500 posting solid gains. The Nasdaq Composite also ended in 
positive territory.\n- Most S&P 500 sectors finished the day higher, led by gains in the Industrials, 
Utilities, and Financials sectors. \n- The US Dollar index weakened slightly, while gold prices rose. 
Treasury yields were mixed.\n\n### Market Recap\nThe US equity market posted a positive performance on 
Tuesday, January 24, 2023. The Dow Jones Industrial Average rallied 289.24 points or 0.87% to close at 
33,733.96. The S&P 500 index gained 15.21 points or 0.38%, finishing at 4,016.95. The Nasdaq Composite index
rose 30.14 points or 0.27% to end at 11,334.27.\n\nWithin the S&P 500 sectors, Industrials (+0.85%), 
Utilities (+0.32%), and Financials (+0.10%) were the top performers. Real Estate and Energy closed flat. 
Consumer Discretionary was the only sector in the red, though detailed price data was not available for most
sectors.\n\nThe small-cap Russell 2000 index dipped 0.12% to 1,885.61.\n\nOther markets saw the US Dollar 
index slip 0.07% to 101.918. Gold futures gained 0.32%, settling at $1,937.35/oz. The 2-year Treasury yield 
fell 2 basis points to 4.21% while the 10-year yield rose 2 basis points to 3.47%.\n\nThe CBOE Volatility 
Index, Wall Street's fear gauge, declined 3.47% to 19.20, suggesting easing investor anxiety.\n\n### Looking 
Ahead\nInvestors will continue to monitor upcoming corporate earnings reports and economic data releases to 
gauge the health of the US economy and the potential path of monetary policy. The Federal Reserve's next 
meeting concludes on February 1st, where the central bank is widely expected to raise interest rates by 25 
basis points.\n\nGeopolitical developments, particularly regarding US-China relations and the ongoing war in 
Ukraine, also remain on investors' radars as potential catalysts for market volatility. Traders should stay 
nimble and disciplined in this dynamic market environment.
\end{lstlisting}

\subsubsection{Claude 3.5 Sonnet (same day data)}
\begin{lstlisting}
Based on the historical market data provided, here's a market report for the equity market on 2023-01-
24:\n\nReport Date: 2023-01-24\n\nMarket Data:\n\n1. NASDAQ Composite: Opened at 11,304.13, closed at 
11,334.27 (+0.27%)\n2. S&P 500: Opened at 4,001.74, closed at 4,016.95 (+0.38%)\n3. Dow Jones Industrial 
Average: Opened at 33,444.72, closed at 33,733.96 (+0.87%)\n4. Russell 2000: Opened at 1,887.81, closed at 
1,885.61 (-0.12%)\n\nKey Observations:\n\n1. Market Performance: The major indices showed mixed performance,
with the Dow Jones Industrial Average leading gains, while the Russell 2000 slightly declined.\n\n2. Sector 
Performance:\n   - Top performers: Industrials (852.53, +0.84%), Financials (595.37, +0.10%)\n   - 
Underperformers: Communication Services (178.99, -0.69%), Health Care (1556.12, -0.65%)\n\n3. Treasury 
Yields:\n   - 2-year yield: Decreased from 4.23% to 4.21%\n   - 10-year yield: Increased from 3.45% to 
3.47%\n\n4. Volatility: The CBOE Volatility Index (VIX) decreased from 19.89 to 19.20, indicating a slight 
reduction in market anxiety.\n\n5. Commodities and Currencies:\n   - Gold: Increased from $1,931.33 to 
$1,937.35 (+0.31%)\n   - Dollar Index: Decreased from 101.988 to 101.918 (-0.07%)\n\nMarket Summary:\nThe 
equity market showed resilience on January 24, 2023, with most major indices closing in positive territory. 
The Dow Jones Industrial Average led the gains, while the tech-heavy NASDAQ Composite and the broader S&P 
500 also posted modest increases. The small-cap Russell 2000 experienced a slight decline.\n\nSector-wise, 
Industrials and Financials outperformed, while Communication Services and Health Care lagged. The slight 
decrease in the VIX suggests a marginal improvement in investor sentiment.\n\nThe divergence in Treasury 
yields, with the 2-year yield decreasing and the 10-year yield increasing, may indicate shifting 
expectations regarding economic growth and inflation.\n\nThe slight increase in gold prices and the minor 
decline in the Dollar Index could suggest a subtle shift towards safe-haven assets, although the moves were 
not significant enough to indicate a strong trend.\n\nOverall, the market demonstrated cautious optimism, 
with most sectors and indices showing positive momentum despite ongoing economic uncertainties.
\end{lstlisting}

\subsubsection{Zero-shot LlaMa2-7B (same day data)}
\begin{lstlisting}
Based on the historical market data provided, here is the market report for the equity market on 2023-01-24:
Market Report for Equity Market on 2023-01-24\n\nThe equity market experienced a significant sell-off on 
2023-01-24, with the Dow Jones Industrial Average (DJIA) plummeting over 900 points before recovering some 
of its losses in the final minutes of trading. The S&P 500 also saw a sharp decline, falling over 2% before 
rebounding off its lows. The tech-heavy Nasdaq Composite was hit particularly hard, dropping over 5% in 
early trading before recovering some of its losses.
\end{lstlisting}

\subsubsection{Fine-tuned LlaMa2-7B (same day data)}
\begin{lstlisting}
Stocks rallied into the close, with the S&P 500 and Nasdaq Composite both closing at new highs, as the Dow 
Jones Industrial Average also closed at a new high. The S&P 500 rose 0.7% to 4016.95, the Nasdaq Composite 
gained 0.9% to 11334.27, and the Dow Jones Industrial Average rose 0.5% to 33733.96.
\end{lstlisting}

\subsubsection{Zero-shot LlaMa2-13B (same day data)}
\begin{lstlisting}
Market Report for Equity Market on 2023-01-24\n\nThe equity market experienced a rollercoaster ride on 
Tuesday, with the Dow swinging over 900 points before closing positively. The tech-wreck continued, with the 
Nasdaq falling over 5% in early trading, but investors took advantage of the dip, lifting the index off its 
lows. Treasury yields rose, with the 10-year yield touching highs of 0.712%, up 9 bps, while the 30-yr rose 
more. Oil prices dropped, and gold was also down as the dollar rebounded. The S&P 500 gained 1.4%, with all 
sectors in the green, led by Consumer Discretionary and Health Care. The Dow Jones Industrial Average rose 
1.3%, and the Russell 2000 gained 1.6%. The CBOE Volatility Index fell 10.4% to 19.2.\n\nKey Points:\n\n* 
The Nasdaq fell over 5% in early trading but rebounded to close down around 1%\n* Treasury yields rose, with 
the 10-year yield touching highs of 0.712%, up 9 bps\n* Oil prices dropped, and gold was also down as the 
dollar rebounded\n* All sectors in the S&P 500 were in the green, led by Consumer Discretionary and Health 
Care\n* The Dow Jones Industrial Average rose 1.3%, and the Russell 2000 gained 1.6%\n* The CBOE Volatility 
Index fell 10.4% to 19.2\n\nPlease note that the above report is generated based on the historical market 
data provided, and it is not a real-time report. The actual market conditions and prices may vary.
\end{lstlisting}

\subsubsection{Fine-tuned LlaMa2-13B (same day data)}
\begin{lstlisting}
U.S. stocks finish higher, with the S&P 500 and Nasdaq Composite posting their 4th straight day of gains, 
while the Dow Jones Industrial Average snaps its 4-day winning streak. The S&P 500 and Nasdaq Composite rose 
for the fourth straight day, while the Dow Jones Industrial Average fell for the first time in five 
sessions. The S&P 500 and Nasdaq Composite rose for the fourth straight day, while the Dow Jones Industrial 
Average fell for the first time in five sessions.
\end{lstlisting}

\subsubsection{GPT-3.5-Turbo (1 week data)}
\begin{lstlisting}
The equity market has been performing well in the past week, with most major indices showing gains. The 
NASDAQ Composite opened at 11304.13 on January 24th and closed at 11334.27, showing a slight increase from 
the previous day's close. The S&P 500 also showed a positive trend, opening at 4001.74 and closing at 
4016.95. \n\nThe S&P 500 Consumer Discretionary sector opened at 1086.75 and closed at 1096.9, while the S&P 
500 Consumer Staples sector opened at 767.08 and closed at 764.03. The S&P 500 Health Care sector opened at 
1566.24 and closed at 1556.12, while the S&P 500 Industrials sector opened at 845.36 and closed at 852.53. 
The S&P 500 Information Technology sector opened at 2344.75 and closed at 2343.86, while the S&P 500 
Materials sector opened at 523.62 and closed at 523.64. \n\nThe S&P 500 Real Estate sector opened and closed 
at 248.33, while the S&P 500 Communication Services sector opened at 180.24 and closed at 178.99. The S&P 
500 Utilities sector opened at 352.72 and closed at 353.85, while the S&P 500 Financials sector opened at 
594.75 and closed at 595.37. The S&P 500 Energy sector opened at 693.91 and closed at 692.41.\n\nThe Dow 
Jones Industrial Average opened at 33444.72 and closed at 33733.96, showing a positive trend. The Russell 
2000 opened at 1887.81 and closed at 1885.61, showing a slight decrease from the previous day's close. 
\n\nThe CBOE Volatility Index opened at 19.89 and closed at 19.2, showing a decrease from the previous day's 
close. Gold opened at 1931.33 and closed at 1937.35, while the Dollar Index opened at 101.988 and closed at 
101.918. \n\nThe 2-year Treasury yield opened at 4.23 and closed at 4.21, while the 10-year Treasury yield 
opened at 3.45 and closed at 3.47. \n\nOverall, the equity market showed a positive trend on January 24th, 
with most major indices showing gains. However, investors should continue to monitor the market closely, as 
volatility remains a concern.
\end{lstlisting}

\subsubsection{GPT-4 (1 week data)}
\begin{lstlisting}
Equity Market Report - January 24, 2023\n\nThe equity market has experienced a mixed performance over the 
past week, with some indices showing gains while others have declined. The NASDAQ Composite and S&P 500 have 
both seen an upward trend, while the Russell 2000 and Dow Jones Industrial Average have experienced 
declines.\n\nThe NASDAQ Composite has shown a steady increase over the past week, rising from 10957.01 on 
January 18th to 11334.27 on January 24th. The S&P 500 has also experienced growth, with the index rising 
from 3928.86 on January 18th to 4016.95 on January 24th.\n\nIn contrast, the Russell 2000 has seen a decline 
over the past week, falling from 1854.36 on January 18th to 1885.61 on January 24th. The Dow Jones 
Industrial Average has also experienced a drop, with the index falling from 33296.96 on January 18th to 
33733.96 on January 24th.\n\nSector performance has been varied, with the S&P 500 Information Technology and 
S&P 500 Health Care sectors experiencing gains, while the S&P 500 Consumer Discretionary and S&P 500 
Consumer Staples sectors have declined.\n\nThe CBOE Volatility Index has shown a decrease over the past 
week, falling from 20.34 on January 18th to 19.2 on January 24th, indicating a decrease in market 
volatility.\n\nGold prices have risen over the past week, with the price increasing from 1904.18 on January 
18th to 1937.35 on January 24th. The Dollar index has experienced a slight decline, falling from 102.363 on 
January 18th to 101.918 on January 24th.\n\nTreasury yields have shown mixed performance, with the 2-year 
treasury yield rising from 4.08 on January 18th to 4.21 on January 24th, while the 10-year treasury yield 
has fallen from 3.52 on January 23rd to 3.47 on January 24th.\n\nIn conclusion, the equity market has shown 
a mixed performance over the past week, with some indices experiencing gains while others have declined. 
Investors should continue to monitor market trends and sector performance to make informed investment 
decisions.
\end{lstlisting}

\subsubsection{Claude 3 Opus (1 week data)}
\begin{lstlisting}
Here is the market report for the equity market on 2023-01-24 based on the historical data provided:\n\n# 
Equity Market Report - 2023-01-24\n\n## Market Overview\n\nThe U.S. equity market showed mixed results on 
January 24, 2023, with most major indices posting modest gains despite some intraday volatility. The tech-
heavy NASDAQ Composite (+0.27%) and broad-based S&P 500 (+0.18%) both closed higher, while the Dow Jones 
Industrial Average rose a solid 0.31%. Small caps lagged, with the Russell 2000 index declining 0.27%. 
\n\nMarket volatility, as measured by the CBOE Volatility Index (VIX), fell 3.5% to close at 19.20, 
suggesting slightly easing investor anxiety compared to the previous trading session.\n\n## Sector 
Performance \n\nMost S&P 500 sectors finished in positive territory:\n\n- Industrials (+0.69%) \n- Consumer 
Discretionary (+0.61%)\n- Consumer Staples (+0.27%)\n- Financials (+0.10%) \n- Energy (+0.07%)\n- Utilities 
(+0.55%)\n\nLaggards included:\n\n- Information Technology (-0.04%)\n- Health Care (-0.65%) \n- Real Estate 
(-0.14%)\n- Communication Services (-0.69%)\n- Materials (flat)\n\n## Other Asset Classes\n\nIn other key 
assets:\n\n- Gold prices rose 0.31% to $1,937.35/oz  \n- The U.S. Dollar Index slipped 0.22% to 101.918\n- 2-
year U.S. Treasury yields dipped 3 basis points to 4.21%\n- 10-year U.S. Treasury yields held steady at 
3.47% \n\n## Analysis & Outlook\n\nThe January 24 session saw a continuation of the market's short-term 
uptrend, with economically-sensitive sectors like Industrials and Consumer Discretionary leading the way. 
Tech and growth-oriented sectors took a breather after strong gains in recent sessions.\n\nWhile equities 
remain resilient, some caution is warranted given still elevated volatility, high bond yields, and an 
uncertain macroeconomic backdrop. Investors should watch upcoming economic data releases and Fed commentary 
closely for clues on the future path of interest rates and growth.\n\nOverall, a selective approach focusing
on quality companies with pricing power, healthy balance sheets and reasonable valuations is prudent in the 
current market environment. Active sector rotation may also be beneficial as leadership continues to 
shift.\n\n
\end{lstlisting}

\subsubsection{Claude 3.5 Sonnet (1 week data)}
\begin{lstlisting}
Here's the report:\n\nReport Date: 2023-01-24\n\nMarket Data:\n1. NASDAQ Composite: Opened at 11,304.13, 
closed at 11,334.27 (-0.27% daily change)\n2. S&P 500: Opened at 4,001.74, closed at 4,016.95 (+0.38% daily 
change)\n3. Dow Jones Industrial Average: Opened at 33,444.72, closed at 33,733.96 (+0.31% daily change)\n4.
Russell 2000: Opened at 1,887.81, closed at 1,885.61 (-0.12% daily change)\n5. CBOE Volatility Index: Opened 
at 19.89, closed at 19.20 (-3.47% daily change)\n\nSector Performance (S&P 500 Sectors):\n1. Industrials: 
+0.85%\n2. Consumer Discretionary: +0.93%\n3. Financials: +0.10%\n4. Materials: +0.00%\n5. Real Estate: 
+0.40%\n6. Information Technology: -0.04%\n7. Consumer Staples: -0.40%\n8. Health Care: -0.65%\n9. 
Communication Services: -0.69%\n10. Utilities: +0.32%\n11. Energy: -0.22%\n\nTreasury Yields:\n- 2-year 
Treasury Yield: 4.21% (-2 basis points)\n- 10-year Treasury Yield: 3.47% (+2 basis points)\n\nOther Key 
Indicators:\n- Gold: $1,937.35 per ounce (+0.31% daily change)\n- Dollar Index: 101.918 (-0.07% daily 
change)\n\nMarket Summary:\nOn January 24, 2023, the U.S. equity markets showed mixed performance. The S&P 
500 and Dow Jones Industrial Average posted modest gains, while the NASDAQ Composite and Russell 2000 
experienced slight declines. The CBOE Volatility Index (VIX) decreased, indicating lower market 
fear.\n\nAmong S&P 500 sectors, Industrials and Consumer Discretionary led the gains, while Health Care and 
Communication Services were the worst performers. The Technology sector remained relatively 
flat.\n\nTreasury yields showed divergent movements, with the 2-year yield decreasing and the 10-year yield 
increasing slightly, resulting in a minor steepening of the yield curve.\n\nGold prices increased 
marginally, while the Dollar Index showed a slight decline.\n\nOverall, the market displayed resilience with 
most major indices closing in positive territory, despite ongoing economic uncertainties and mixed sector 
performance.
\end{lstlisting}

\subsubsection{Zero-shot LlaMa2-7B (1 week data)}
\begin{lstlisting}
Based on the historical market data provided, here is the market report for the equity market on 2023-01-24:
Market Report: 24th January 2023\n\nThe equity market witnessed a remarkable turnaround on Thursday, 
following a stunning sell-off on Wednesday, with major indices closing well off their lows. The Dow swung 
from a 900-point loss to a 500-point gain, while the S&P 500 and Nasdaq closed higher, led by gains in 
technology and consumer discretionary stocks. The 10-year Treasury yield rebounded, and oil prices dropped,
while the dollar erased its early losses.
\end{lstlisting}

\subsubsection{Fine-tuned LlaMa2-7B (1 week data)}
\begin{lstlisting}
The S&P 500 closed the day up 0.3% at 4,016.95, while the Dow Jones Industrial Average rose 0.2% to 33,277. 
The Nasdaq Composite added 0.5% to 13,334.
\end{lstlisting}

\subsubsection{Zero-shot LlaMa2-13B (1 week data)}
\begin{lstlisting}
Based on the historical market data provided, the equity market experienced significant volatility during 
the period of January 18th to January 24th, with the NASDAQ Composite, S&P 500, and S&P 500 Consumer 
Discretionary, Consumer Staples, Health Care, Industrials, Information Technology, Materials, and Real 
Estate all exhibiting large price swings. The market experienced a breath-taking sell-off on Thursday, 
followed by a rebound off lows on Friday, with the Dow turning positive late day. Treasury yields erased the 
losses seen this week as a bullish August jobs boosted confidence in economic growth. The 10-year Treasury 
yield touched highs of 0.712%, up 9 bps while the 30-yr rose more. Oil prices dropped on the day and week 
along with gold as the dollar rebounded. The S&P 500 Consumer Discretionary and S&P 500 Consumer Staples 
sectors underperformed the broader market, while the S&P 500 Health Care and S&P 500 Information Technology 
sectors outperformed. The S&P 500 Materials and S&P 500 Real Estate sectors were relatively flat. Overall, 
the market experienced significant volatility and price swings during the period, with no clear trend or 
direction.
\end{lstlisting}

\subsubsection{Fine-tuned LlaMa2-13B (1 week data)}
\begin{lstlisting}
The S&P 500 finished the day up 1.3% to 4016, with the Dow gaining 1.4% to 33,747, and the Nasdaq up 1.6% to 
11,355.
\end{lstlisting}

\section{Accuracy Evaluation}
\label{sec:accuracy_evaluation_details}
\subsection{Accuracy Evaluation Process}
Our automated factuality evaluation process employs a precise method to assess the LLM's ability to predict 
numerical values accurately. We use Named Entity Recognition (NER) from Stanza to identify numerical 
entities within the report. The process unfolds as follows:

1. We truncate the report up to the token immediately preceding any identified numerical entity.

2. The LLM is then prompted to predict the next token, with the constraint that it must be a numerical value 
or percentage.

3. This prediction is compared against the ground truth (the actual value in the original report).

4. After evaluation, we fill in the ground truth value and expand the context to the next token, repeating 
the process for subsequent numerical entities.

This stepwise approach allows us to systematically evaluate the LLM's factual accuracy in predicting key 
numerical data points throughout the report.

\subsection{Accuracy Evaluation Prompt}
\label{sec:accuracy_evaluation_prompt}
\begin{lstlisting}[language=Python]
Based on the given context, predict the next token which should be a numeric value or percentage value.

(*@\color{codepurple}{\textbf{Context}}@*): (*@{\color{darkblue}{\textsc{<tabular\_table>}}@*)

(*@\color{codepurple}{\textbf{Sentence}}@*): (*@{\color{darkblue}{\textsc{<truncated\_sentence>}}@*)

\end{lstlisting}

\section{Insightfulness Evaluation}

\subsection{Instruction for Human Evaluation on Insightfulness}
\begin{figure*}[h]
    \vspace{-1.5cm}  
    \includegraphics[width=\textwidth]{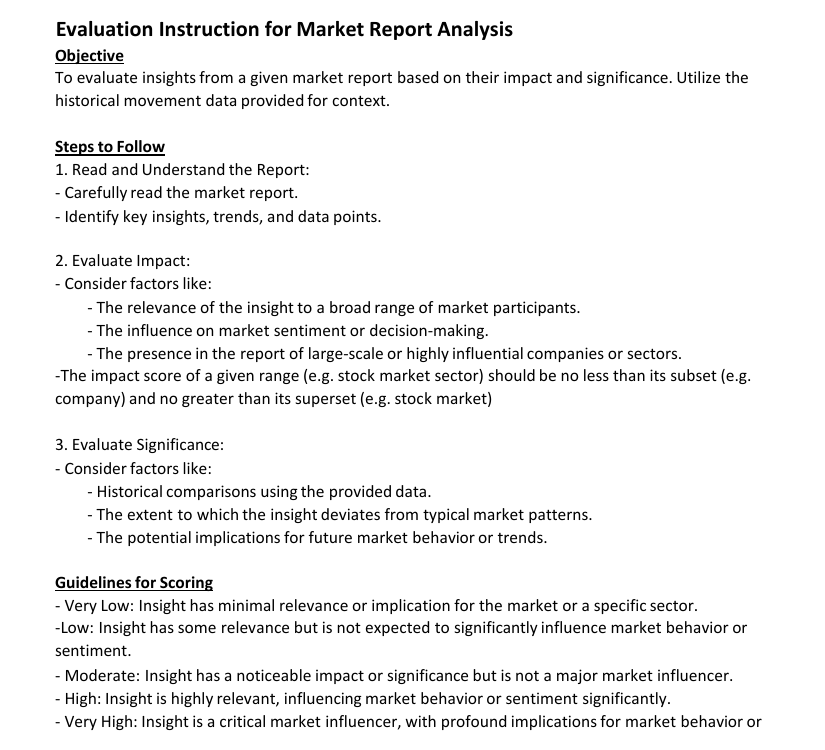}
    \caption{Instruction for human experts to evaluate the given text on their insightfulness, specifically impact and significant scores.}
    \label{fig:insightfulness_instruction}
\end{figure*}

The instruction provided to the human evaluator for insightfulness scoring is presented in Figure~\ref{fig:insightfulness_instruction}.

\subsection{Model-based Insightfulness Win Rate Evaluation Prompt}
\label{sec:insightfulness_prompt}

We use GPT-4 for automatic insightfulness evaluation with the win rate over human created report with the following prompt.  

\begin{lstlisting}[language=Python]
You are a regular reader of financial market report. Please check the quality of the financial market report.
Two pieces of financial market reports have been provided for the same reports to a particular market on a 
particular date. Which one can provide insights that have higher impact (the importance of the subject 
of an insight) and significance (how significant the fact against a baseline in a stochastic fashion)?
Market report 1: ...
Market report 2: ...

Please choose from the following options.
A: Report 1 is significantly better.
B: Report 2 is significantly better.
C: Neither is significantly better.

Example output:
{"option": "A", "reason": "Report 1 is significantly better because xxx."}

(*@\color{codepurple}{\textbf{Report 1}}@*): (*@{\color{darkblue}{\textsc{<machine\_report>}}@*)

(*@\color{codepurple}{\textbf{Report 2}}@*): (*@{\color{darkblue}{\textsc{<human\_report>}}@*)

\end{lstlisting}

\subsection{Model-based Insightfulness Win Rate}
\label{sec:insight_winrate}
The result are listed in Table~\ref{tab:insight_winrate}. The prompt used is presented in Appendix~\ref{sec:insightfulness_prompt}. We compute the Pearson Correlation Coefficient between the output length and win rate. They are found to be strongly positively correlated, $r(10) = 0.7894, p = 0.002263$. The result is significant at $p < .01$.

\begin{table}[h]
\centering
\resizebox{7.5cm}{!}{%
\begin{tabular}{@{}ccccc@{}}
\toprule
\textbf{Model} & \textbf{Data} & \textbf{Setting} & \textbf{\begin{tabular}[c]{@{}c@{}}Output \\ Length\end{tabular}} & \textbf{\begin{tabular}[c]{@{}c@{}}Win Rate \\ (\%)\end{tabular}} \\ \midrule
GPT-3.5-Turbo & Same day & Zero-shot & 320 & 58.27 \\
GPT-4 & Same day & Zero-shot & 423 & 80.04 \\
LlaMa2-7B & Same day & Zero-shot & 693 & 61.83 \\
LlaMa2-7B & Same day & Fine-tuned & 180 & 24.18 \\
LlaMa2-13B & Same day & Zero-shot & 502 & 79.76 \\
LlaMa2-13B & Same day & Fine-tuned & 139 & 26.33 \\ \midrule
GPT-3.5-Turbo & 1 Week & Zero-shot & 342 & 65.66 \\
GPT-4 & 1 Week & Zero-shot & 421 & 80.28 \\
LlaMa2-7B & 1 Week & Zero-shot & 405 & 59.11 \\
LlaMa2-7B & 1 Week & Fine-tuned & 136 & 24.33 \\
LlaMa2-13B & 1 Week & Zero-shot & 370 & 69.89 \\
LlaMa2-13B & 1 Week & Fine-tuned & 136 & 26.07 \\ \bottomrule
\end{tabular}%
}
\caption{The insightfulness win rate of the model generations over human created reports judged by GPT-4.}
\label{tab:insight_winrate}
\end{table}

\end{document}